%% file: mian.tex
\pdfoutput=1
\documentclass[10pt,twocolumn,letterpaper]{article}

\usepackage{cvpr}              

\input{preamble}

\definecolor{cvprblue}{rgb}{0.21,0.49,0.74}
\usepackage[pagebackref,breaklinks,colorlinks,citecolor=cvprblue]{hyperref}

\newcommand{\cb}[1]{\textcolor[rgb]{0,0,0}{#1}}

\usepackage[normalem]{ulem}
\useunder{\uline}{\ul}{}
\usepackage{multirow}
\usepackage{colortbl}

\def\paperID{3944} 
\def\confName{CVPR}
\def\confYear{2024}

\title{Task-Customized Mixture of Adapters for General Image Fusion}

\author{Pengfei Zhu, Yang Sun, Bing Cao\thanks{Corresponding author.} , Qinghua Hu\\
Tianjin Key Lab of Machine Learning, College of Intelligence and Computing, Tianjin University, China\\
\tt\small {
\{zhupengfei,yangsun,caobing,huqinghua\}@tju.edu.cn
}
}

\begin{document}
\maketitle

\input{arxiv/sec/0_abstract}    
\input{arxiv/sec/1_intro}
\input{arxiv/sec/2_RelatedWork}
\input{arxiv/sec/3_Method}
\input{arxiv/sec/4_Experimental_Results}
\input{arxiv/sec/5_Conclusion}
\input{arxiv/sec/6_Acknowledgement}

{
    \small
    \bibliographystyle{ieeenat_fullname}
    \bibliography{main}
}
\input{arxiv/sec/X_suppl}



\end{document}

%% file: preamble.tex
\usepackage[dvipsnames]{xcolor}


%% file: arxiv/sec/0_abstract.tex
\begin{abstract}
\cb{General image fusion aims at integrating important information from multi-source images. However, due to the significant cross-task gap, the respective fusion mechanism varies considerably in practice, resulting in limited performance across subtasks. To handle this problem, we propose a novel task-customized mixture of adapters (TC-MoA) for general image fusion, adaptively prompting various fusion tasks in a unified model. We borrow the insight from the mixture of experts (MoE), taking the experts as efficient tuning adapters to prompt a pre-trained foundation model. 
These adapters are shared across different tasks and constrained by mutual information regularization, ensuring compatibility with different tasks while complementarity for multi-source images. The task-specific routing networks customize these adapters to extract task-specific information from different sources with dynamic dominant intensity, performing adaptive visual feature prompt fusion. Notably, our TC-MoA controls the dominant intensity bias for different fusion tasks, successfully unifying multiple fusion tasks in a single model. Extensive experiments show that TC-MoA outperforms the competing approaches in learning commonalities while retaining compatibility for general image fusion (multi-modal, multi-exposure, and multi-focus), and also demonstrating striking controllability on more generalization experiments. The code is available at \url{https://github.com/YangSun22/TC-MoA}.}
\end{abstract}

%% file: arxiv/sec/1_intro.tex
\section{Introduction}
\label{sec:intro}

\begin{figure}
  \centering
  \includegraphics[width=\linewidth]{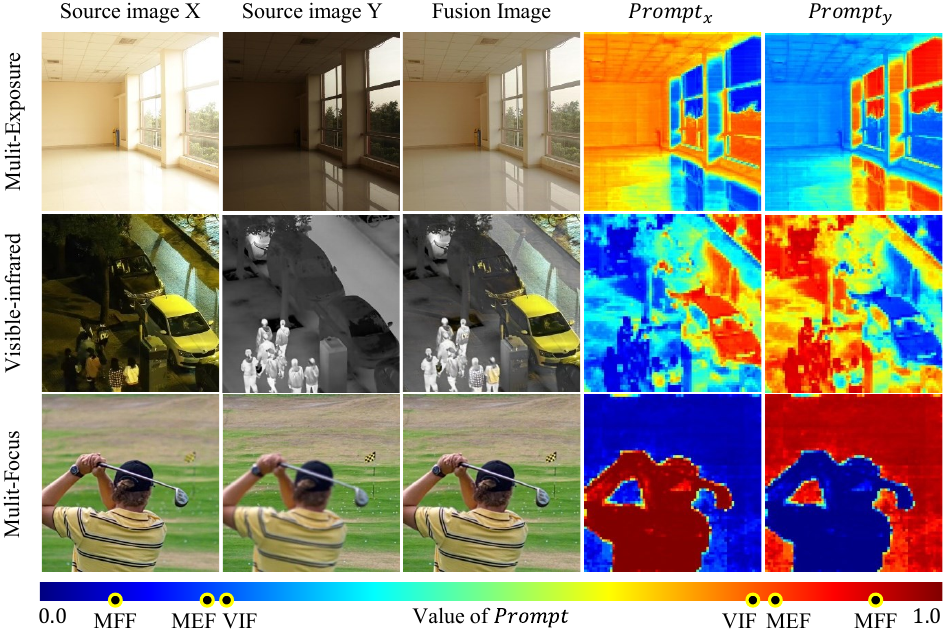}
  \vskip -0.1in
   \caption{Prompt can adaptively select the complementary information from multi-source features. The dominant intensity bias vary according to the task, which is reflected by the different shades of colors.}
   \label{fig:introduction}
    \vspace{-1.8em} 
\end{figure}

Image fusion aims to integrate complementary information from multi-source images captured by different sensors in the same scene onto a single image. It is usually used to enhance important information and visual quality \cite{IF_Survey}. 
Currently, general image fusion mainly includes multi-modal, multi-exposure, and multi-focus image fusion, \etc 
Fusion tasks exhibit diverse fusion mechanisms.
The \textit{Multi-Exposure image Fusion} (MEF) focuses on integrating images with multiple exposure levels into a high-quality full exposure image~\cite{MEF_Survey}. 
Each source image contributes its own illumination and structural information to the fused image~\cite{MEF-Net}. 
The \textit{Visible-Infrared image Fusion} (VIF) is a subfield of the \textit{Multi-Modal Fusion} (MMF)~\cite{DenseFuse,Trusted_Multi-View,zhang2023provable} that aims at fusing complementary information of infrared and visible modalities to produce robust and informative fused images \cite{IVF_Survey}.  Infrared images provide more intensity information, while visible images offer more texture and gradient information. 
The \textit{Multi-Focus image Fusion} (MFF) aims at generating an all-in-focus image from a series of partially focused images \cite{MFF_Survey}. A patch of multi-focus fused image typically corresponds to only one source image. 
Consequently, it can be observed that the MEF and VIF tasks involve relatively balanced fusion, while MFF is an extremely unbalanced task that tends to exhibit polarized choices. 

With the rapid development of deep learning techniques, image fusion has achieved great progress in recent years~\cite{Deepfuse,U2Fusion}, while most existing methods~\cite{MEF-Net,DDFM,CDDFusion} focus on single image fusion scenario alone. 
Task-specific methods~\cite{MEF-Net,YDTR} often employ task-specific strategies such as designing complex task-biased networks or utilizing task-specific fusion loss functions, resulting in the weak generalization to other tasks~\cite{U2Fusion}.
Considering that different fusion tasks have similar objectives, \ie, integrating complementary information from multiple source images, some recently proposed approaches~\cite{IFCNN,U2Fusion,DeFusion} attempted to conduct various fusion tasks using a unified model, forming general image fusion.
These methods, however, are either enmeshed in dominant-task bias~\cite{IFCNN} or multi-task commonality~\cite{U2Fusion,DeFusion} at the expense of individuality, leading to unsatisfactory performance. 
This motivates us to explore a more compatible fusion paradigm that is adaptively and dynamically compatible with different fusion scenarios.

To handle this challenge, inspired by the impressive representation capability of pre-trained foundation models, we introduce the foundation model as a fixed encoder to extract complementary features for multi-source images.
Different from most existing methods, we borrow insight from the mixture of experts (MoE)~\cite{MoE,MMoE}, taking each expert as an efficient tuning adapter to perform adaptive visual feature prompt fusion based on the foundation model.
The task-specific routing networks customize these adapters to generate task-specific fusion prompt for different sources, forming a novel task-customized mixture of adapters (TC-MoA) architecture.
An mutual information regularization is further developed to constrain the prompt, which guarantees complementarity for diverse sources.
It is worth noting that the prompt have significant task bias and dominant intensity gaps. As shown in \cref{fig:introduction}, the prompt of MFF has a greater color difference than VIF and MEF, meaning that feature selection has more bipolarity on dominant intensity bias. Our model effectively perceives the fusion intensity bias among varied fusion tasks in a single model and is therefore compatible with a wider range of fusion tasks.
Extensive experiments verify our superiority over the competing methods in general image fusion, including multi-modal, multi-exposure, and multi-focus fusion.
More importantly, our TC-MoA even shows unprecedented controllability and generalizability for unseen fusion tasks, fully demonstrating our potential in broader scenarios. We summarize our main contributions as follows:

\begin{itemize}
\item We propose a unified general image fusion model, providing a new task-customized mixture of adapters (TC-MoA) for adaptive multi-source image fusion
(benefiting from the dynamically aggregating effective information from the respective modalities).
\item We propose a mutual information regularization for adapters, which allows our model to more accurately identify the dominant intensity of different source images.
\item To the best of our knowledge, we for the first time pro-
pose an MoE-based flexible adapter for the foundation model in general image fusion. By only adding 2.8\% of learnable parameters, our model copes with numerous fusion tasks. Extensive experiments demonstrate our superiority against the competing methods, while showing significant controllability and generalizability.
\end{itemize}



%% file: arxiv/sec/2_RelatedWork.tex
\section{Related Work}
\label{sec:formatting}



\noindent\textbf{Image Fusion.}
Image fusion focuses on generating a fused image containing complementary information from different source images.
%
Some early methods \cite{DenseFuse,Deepfuse,CNN} tackled their respective task by leveraging CNN. 
Then, GAN-based \cite{MEF-GAN,MFI-WHU} and Transformer-based 
methods~\cite{YDTR,TransMEF,UTUFuse-Net} have been proposed to improve the fusion quality.
Furthermore, high-level tasks \cite{TarDAL,DetFusion, MoE-Fusion} are also introduced to guide the fusion of images. Feature decomposition \cite{DIDFuse,AUIF} for high-low frequency has also gained significant attention as a research direction. 
Different from these methods, some most recent methods focus on general image fusion that aims to address multiple fusion tasks in a single model.
\citet{IFCNN} performed a supervised training framework for MFF, which was generalized to other tasks by adjusting fusion conditions.
\citet{FusionDN,U2Fusion} appraised the quantity and quality of information from source images or features with the image quality assessment (IQA) to decide the fusion paradigm. 
PMGI \cite{PMGI} and SwinFusion \cite{swinfusion} incorporated a unified fusion framework and loss function, and separate model training for individual tasks.
\citet{DeFusion} performed a self-supervised fusion framework to learn fusion commonality while ignoring task-specific individuality. 
In this work, we accommodate diverse fusion tasks by dynamically customizing the mixture of adapters, rather than suffering individuality for cross-task commonality.

\noindent\textbf{Parameter-Efficient Fine-Tuning.}
To efficiently adapt pre-trained models to the respective downstream application tasks, some \textit{Parameter-Efficient Fine-Tuning} (PEFT) studies have been proposed.
PEFT can be mainly divided into the Adapter \cite{Adapter, LoRA, SSF} and the Prefix Tuning \cite{PrefixTuning, VPT}. 
\citet{VPT} introduced prompt tuning to the ViT structure. \citet{AdaptFormer} proposed a new adapter by scaling the original features instead of summing.
\citet{SSF} bridged the gap between the pre-trained features and the downstream task features by linearly varying the original features. 
In this paper, different from these methods that focus on tuning high-level tasks, we for the first time generalize the powerful foundational model to general image fusion.

\noindent\textbf{Mixture of Experts.}
\citet{MoE} first proposed the mixture of experts (MoE) to increase model capacity without increasing computational complexity. Based on this, \citet{GShard} and \citet{Switch} integrated MoE with the transformer structure, further pushing the upper limit of network capacity. In addition, MoE has been validated as effective in other challenges. 
\citet{MMoE} tackled the multi-task problem by designing multi-gates MoE. 
\citet{LIMoE} exploited MoE to train the contrastive learning-based vision-language foundation model. 
Inspired by this, we take each expert as an adapter, forming a task-customized mixture of adapters for tuning our general image fusion framework. 

%% file: arxiv/sec/3_Method.tex
\begin{figure*}[t]
  \centering
    \includegraphics[width=1\linewidth]{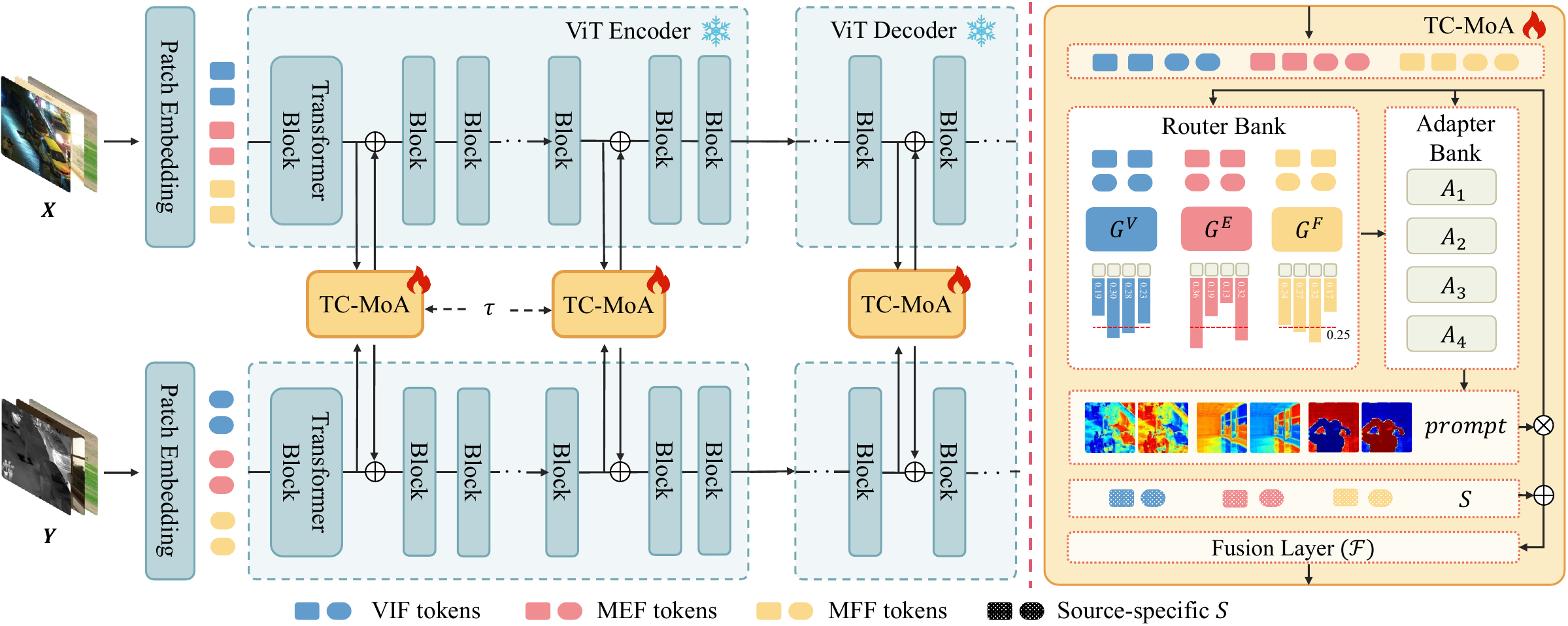}
    \vspace{-1.8em}
    \caption{ An overview of our proposed TC-MoA method. Our approach gradually modulates the fusion results by inserting TC-MoA into the frozen ViT backbone. TC-MoA generates task-specific prompt through a task-specific router bank and an shared adapter bank. The fusion layer utilizes prompt as scale and source-specific embeddings as biases to obtain fusion images.}
    
    \label{fig:method_1}
    \vspace{-1em}
\end{figure*}

\section{Method}



\subsection{Overview}
In this paper, we propose a task-customized mixture of adapters for general image fusion, which is a novel parameter-efficient fine-tuning method to jointly and adaptively prompt various fusion tasks.
Given a pair of source images $ \left \{ (X,Y)  | X,Y\in \mathbb{R}^{H\times W\times 3}\right \}$, the network integrates complementary information from different sources to obtain the fused image $ I_{Fusion} \in \mathbb{R}^{H\times W\times 3} $. As shown in \cref{fig:method_1}, we feed the source images into the ViT network and obtain tokens of the source images through the patch embedding layer. ViT consists of an encoder for feature extraction and a decoder for image reconstruction, both of which are composed of transformer blocks. A TC-MoA is inserted into each $ \tau $ $ (\tau=4)$ transformer blocks in both the encoder and decoder. 
Each TC-MoA consists of a task-specific router bank, a task-shared adapter bank, and a fusion layer. Our network gradually modulates the fusion results through these TC-MoA.


\subsection{Task-Customized Mixture of Adapters}
As shown in \cref{fig:method_1} , each TC-MoA consists of a task-specific router bank $ \left \{ G^{\mathcal{V} }, G^{E}, G^{F} \right \} $, a task-shared adapter bank $  \left \{ A_{1},\dots,A_{N}  \right \}  $  and a prompting fusion layer $  \mathcal{F} $ . The TC-MoA includes two main stages: prompt generation and prompt-driven fusion. For ease of expression, we take the VIF as an example, assuming that the input comes from the VIF dataset and uses $G$ to denote $G^V$.

\noindent\textbf{Prompt Generation. }
First, we obtain multi-source features for subsequent processing. The network structure before the $j$-th TC-MoA is defined as $E_{j}(\cdot)$, and the extracted features are defined as $f_{x}$ and $f_{y}$, where $f \in \mathbb{R}^{ pH\times pW \times C} $. We concatenate $ f_{x}   $ and $f_{y}$ as a feature representation of pairs of multi-source tokens. 
This allows tokens from different sources to exchange information in the subsequent network. However, directly computing on the high-dimensional concatenated features will bring amounts of unnecessary parameters. Therefore, we use $L(\cdot)$ to perform feature dimension reduction and obtain the processed multiple-source features $\Phi$ as follows,
\begin{equation}
  f_{x}  = E_{j}(X ),f_{y}  = E_{j}(Y )
  \label{eq:GetFeature_1}
\end{equation}
\begin{equation}
  \Phi   = L(Cat(f_{x} ,f_{y} ))
  \label{eq:GetFeature_2}
\end{equation}
where $Cat(\cdot)$ represents the concatenation of features, $ L(\cdot) $ consists of a linear layer and normalization layers. 
Next, depending on the task to which $ \Phi $ belongs, we select a task-specific router from the router bank to customize the routing schemes, i.e., which adapters in the adapter bank should be input for each token pair.
\begin{multline}
  G(x) =Softmax(TopK(x\cdot W_{g} + \\
  \mathcal{N}(0,1)\cdot Softplus(x\cdot W_{noise} )))
  \label{eq:G_x}
\end{multline}
where $ TopK(\cdot) $ keeps only the top $  K (K = 2) $ values, setting the rest to $ -\infty $ (after $Softmax(\cdot)$, the value becomes 0 ). $  W_{g}  $ and $ W_{noise}   $ are learnable parameters. The customization routing schemes vary for tasks, as evidenced by the proportion of the number of times that different adapters have been routed in \cref{fig:method_1} (b).
After that 
, we weight sum the output from the adapters to obtain the prompt. Each router masters a task-biased appetite for customizing a suitable mixture of adapters, and each adapter generates the prompt. 
The task-customized multi-source prompt is then calculated as follows,
\begin{equation}
  prompt = GAP(Sigmoid(\sum_{i=1}^{N} G(\Phi )_{i}  \cdot A_{i}(\Phi ))
  \label{eq:prompt}
\end{equation}
where $ GAP(\cdot) $ represents the global average pooling, $ i $ is the index of adapters and $ G(\cdot )_{i} $ is the routing value for the $ i$-th  adapter. The task-customized prompt, denoted as $prompt\in \mathbb{R}^{pH\times pW \times2}$, is composed of $prompt_{x}$ and $prompt_{y}\in \mathbb{R}^{pH\times pW \times1}$ and has a value range of $(0,1)$ . 

\noindent\textbf{Prompt-Driven Fusion.}
The task-customized prompt is constrained by mutual information regularization (MIR), which guarantees complementarity for diverse sources. Thus the prompt can be taken as an evaluation of the proportion of important information in each source. 
By dot-multiplication of multi-source features and prompt, we retain complementary information while removing redundant information. Afterward, considering that the feature representation should contain a source-correlated bias
(e.g. visible or infrared images), we introduce input-independent learnable parameters for each source, i.e., source embedding $ S $. After feature prompting and shifting, we obtain the refined source features $ (h_{x},h_{y}) $ as follows,
\begin{equation}
\begin{split}
  h_{x}  =prompt_{x}\cdot f_{x}   + S_{x} \\
 h_{y}  =prompt_{y}\cdot f_{y}   + S_{y} \\
 f_{TC-MoA}=\mathcal{F} (h_{x} +h_{y} )
  \label{eq:prompt_scale}
  \end{split}
\end{equation}
where $S_{x} $ and $S_{y} $ represent the source embeddings of visible and infrared images in VIF respectively.  $\mathcal{F}(\cdot)$ fuses the refined multi-source features using an additive operation and then passes them through a set of convolutional layers. These layers introduce local receptive fields to reduce the checkerboard artifacts and align the solution space of the subsequent transformer blocks. Ultimately, we obtain a fusion feature through the task-customized prompt. To encourage the model to extract important information gradually, the features input into the next transformer block are processed as follows,
\begin{equation}
 \begin{split}
  { f_{x} }' = \lambda _{f}f_{x}  +(1-\lambda _{f}) f_{TC-MoA} \\
  { f_{y} }' = \lambda _{f}f_{y}  +(1-\lambda _{f}) f_{TC-MoA}
 \end{split}
\end{equation}
where $\lambda_{f}$ is a learnable parameter initialized to 0.5.

\noindent\textbf{Mutual Information Regularization.} 
In order to ensure that the model dynamically retains complementary information while discarding redundant information of multi-source features, we impose regularization constraints on the prompt. Assuming that the information representation of features varies linearly, we define the MIR as follows,
\begin{equation}
    \min {\left | prompt_{x}+prompt_{y}-1  \right |} 
  \label{eq:MIR}
\end{equation}
The MIR allows the model to accurately identify the dominant intensity of different sources, which is positively correlated with the information content of the sources.

\subsection{Task-Customized Loss Function}


In addition to accommodating the individuality of different tasks in our network structure, we also customize unsupervised loss functions for each fusion task. We add $ \mathcal{L} _{aux} $ to the loss function of each task to ensure the training of TC-MoA and $ \mathcal{L} _{aux} $ is the auxiliary loss in \cite{MoE} to avoid adapters learning unbalanced. On the other hand, in order to generate high-quality fusion images, we impose constraints on the structural information ($\mathcal{L} _{ssim}$), intensity information ($\mathcal{L} _{Pixel}$) for VIF loss function, and gradient information ($\mathcal{L} _{Grad}$) of the fusion images for different fusion tasks.

For the VIF task, our objective is to retain the most pronounced high-frequency and low-frequency information from the source images in the fused image. Thus, we design the $\mathcal{L} _{MaxPixel}$ and $\mathcal{L} _{MaxGrad}$. To avoid confusing gradients, we retain the sign of the gradient values in all loss functions related to gradient information.
For the MEF task, we consider that the luminance of the fused images should be at an average level with all the gradient information. Thus we design the loss functions for MEF with  $\mathcal{L} _{AvgPixel}$ and ${L} _{MaxGrad}$.  Additionally, we adopt $\mathcal{L} _{mefssim}$~\cite{MEF-Net} which is specially designed for the MEF task, instead of the SSIM loss function.
For the MFF task, we believe that each patch of the fused image should only depend on a single source image with the maximum gradient. This is to prevent the objects' edges in defocused images from being preserved, thereby affecting the quality of the fused image. For this purpose, we select only one of the source images to compute the loss function for each patch in the image, \ie $\mathcal{L} _{MaskPixel}$ and $\mathcal{L} _{MaskGrad}$. Please refer to the supplementary material for details of the loss functions.

%% file: arxiv/sec/4_Experimental_Results.tex
\section{Experiments}
\subsection{Experimental Setting}

\textbf{Datasets.}
We conduct experiments in three image fusion scenarios: visible-infrared, multi-focus, and multi-exposure fusion.
For VIF, we evaluate our model on the LLVIP~\cite{LLVIP} dataset. The training set contains 12025 image pairs, and we randomly select 70 samples out of the test set for evaluation. For MEF, we employ the SCIE \cite{SCIE} dataset (589 pairs) for training and MEFB~\cite{MEF_Benchmark} dataset (100 pairs) for testing, we utilize the most underexposed and overexposed images from sequences in SCIE as inputs.
For MFF, we train our model on the RealMFF~\cite{Real-MFF} and the MFI-WHU~\cite{MFI-WHU} datasets and follow the test setting in MFIF~\cite{MFF_DL_Survey}.
\begin{figure}[t]
  \centering
 
    \includegraphics[width=1\linewidth]{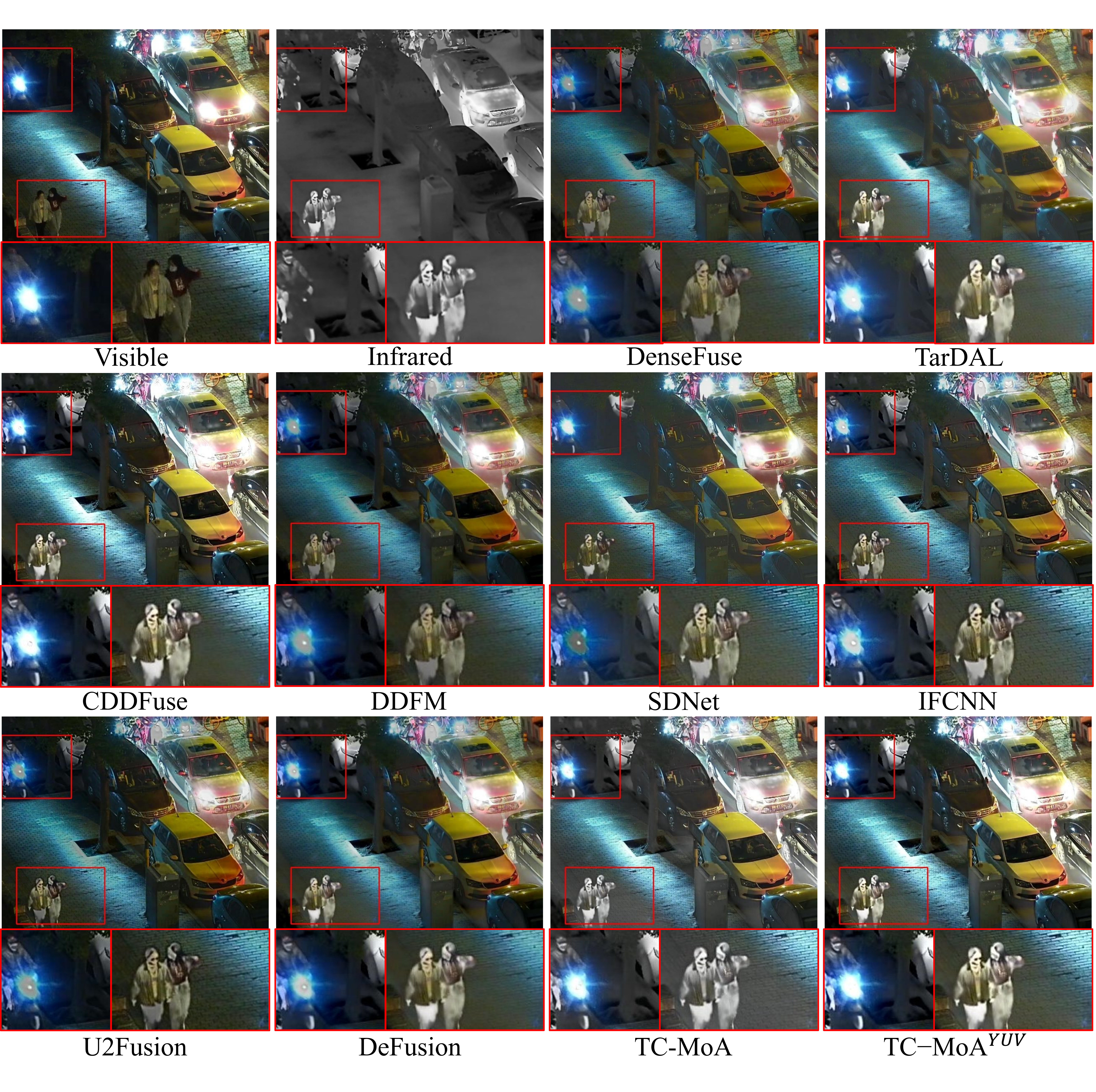}
    \vspace{-1.3em}
    \caption{Qualitative comparisons of various methods in VIF task.}
    \vspace{-1.3em}
    \label{Vis_VIF}
    
\end{figure}

\noindent\textbf{Implementation Details.} 
Our experiments are performed on a server with  8 $\times$ NVIDIA RTX A6000 GPUs. Although our model supports inputs of arbitrary size, we ensure that each fusion task receives an equal amount of data per iteration by randomly cropping all samples to a size of $448\times448$.
We employ the pre-trained MAE-large model~\cite{MAE} with GAN loss as our backbone. We train 20 epochs jointly with all the fusion tasks, and the batch size for each fusion task is set to 3. We adopt the AdamW optimizer with the initial learning rate of $1.5\times10^{-4}$. To ensure training stability, we apply the Exponential Moving Average (EMA) to routers and adapters optimization. Each TC-MoA consists of $ N (N=4) $ adapters embedded in different layers, but only the top $ K (K=2) $ adapters are activated.

\noindent\textbf{Evaluation Metrics.}
Since the commonly used evaluation metrics for fusion tasks are not exactly the same, we customize a set of metrics for each task. Existing metrics can be mainly divided into the following four categories~\cite{MEF_Survey}. $\diamond$ The information theory-based metrics: entropy (EN), peak signal-to-noise ratio (PSNR), mutual information (MI), normalized MI (NMI), Nonlinear correlation information entropy ($ Q_{ncie} $),  feature mutual information (FMI). $\diamond$ The image feature-based metrics: phase congruency-based metric $ Q_{p} $, standard deviation (SD),  gradient-based metric $ Q_{abf} $, and $ Q_{g} $. $\diamond$ The structural similarity-based metrics: $ Q_{c}  $ \cite{Qc},  $Q_{w} $, $Q_{s}$ \cite{Qs}, structural similarity (SSIM), multiscale structural similarity (MS-SSIM \cite{MSSSIM}) and MEF-SSIM \cite{MEFSSIM}. $\diamond$ The human perception inspired metrics: visual information fidelity ($ VIF $), $ Q_{cb} $ and  $Q_{cv} $. Detailed information on metrics can be found in \cite{IVF_Survey} and  \cite{MFF_Survey}.


\begin{table}[t] \scriptsize
  \centering
   \setlength{\tabcolsep}{2.35pt}
   
   \caption{ Quantitative results of the VIF task in LLVIP dataset. \textbf{Boldface in \textcolor[rgb]{ 1,  0,  0}{red}} and \textbf{boldface in \textcolor[rgb]{ 0,  0,  1}{blue}} show the best and second-best values, respectively. The \underline{underline} represents the best value in the general image fusion methods. }
   \vspace{-1.3em}
    \begin{tabular}{@{}lcccccccc@{}}
    \toprule
    Method & $VIF$ & $ Q_{c} $ & EN & SD & \textbf{$Q_{cv}$↓} & MS-SSIM & FMI & $ Q_{w} $  \\
    \midrule
   Densefuse \cite{Deepfuse} & 0.545 & 0.533 & 6.830 & 9.381 & 817.213 & 0.878 & 0.876 & 0.622 \\
    AUIF \cite{AUIF} & 0.402 & 0.370 & 6.137 & 7.769 & 1087.569 & 0.784 & 0.869 & 0.530 \\
    DIDFuse \cite{DIDFuse} & 0.366 & 0.348 & 5.991 & 7.765 & 897.007 & 0.767 & 0.863 & 0.479 \\
    TarDAL \cite{TarDAL}& 0.550 & 0.562 & {\color[HTML]{0000FF} \textbf{7.347}} & 9.609 & 549.177 & 0.864 & 0.860 & 0.628 \\
    YDTR \cite{YDTR} & 0.486 & 0.524 & 6.638 & 8.810 & 883.333 & 0.835 & 0.876 & 0.614 \\
    RFN-Nest \cite{RFN-Nest} & 0.497 & 0.456 & 7.052 & 9.609 & 857.157 & 0.862 & 0.871 & 0.385 \\
    SwinFuse \cite{Swinfuse}& 0.399 & 0.321 & 5.878 & 7.457 & 1306.652 & 0.734 & 0.870 & 0.425 \\
    UMF-CMGR \cite{UMF_CMGR}& 0.414 & 0.479 & 6.607 & 8.520 & 810.670 & 0.801 & 0.876 & 0.561 \\
    DDFM \cite{DDFM}& 0.588 & 0.561 & 7.069 & 9.696 & 760.006 & 0.908 & 0.878 & 0.663 \\ 
    CDDFuse \cite{CDDFusion}& {\color[HTML]{0000FF} \textbf{0.694}} & {\color[HTML]{FF0000} \textbf{0.645}} & 7.342 & {\color[HTML]{0000FF} \textbf{9.733}} & {\color[HTML]{0000FF} \textbf{495.473}} & 0.933 & {\color[HTML]{0000FF} \textbf{0.883}} & 0.830 \\
    \midrule
    SDNet \cite{SDNet}& 0.527 & 0.575 & 6.897 & 9.318 & 936.389 & 0.878 & 0.872 & 0.749 \\
    IFCNN \cite{IFCNN}& 0.679 & 0.634 & 7.223 & 9.662 & 521.741 & {\color[HTML]{0000FF} \textbf{0.946}} & 0.882 & {\color[HTML]{0000FF} \textbf{0.856}} \\
    U2Fusion \cite{U2Fusion}& 0.503 & 0.492 & 6.647 & 8.789 & 857.455 & 0.878 & 0.870 & 0.695 \\
    DeFusion \cite{DeFusion}& 0.606 & 0.606 & 7.216 & 9.701 & 532.092 & 0.890 & 0.880 & 0.676 \\
    TC-MoA & {\color[HTML]{FF0000} {\ul \textbf{0.726}}} & {\color[HTML]{0000FF} {\ul \textbf{0.637}}} & {\color[HTML]{FF0000} {\ul \textbf{7.428}}} & {\color[HTML]{FF0000} {\ul \textbf{9.805}}} & {\color[HTML]{FF0000} {\ul \textbf{423.773}}} & {\color[HTML]{FF0000} {\ul \textbf{0.949}}} & {\color[HTML]{FF0000} {\ul \textbf{0.886}}} & {\color[HTML]{FF0000} {\ul \textbf{0.858}}}
    \\
    \bottomrule

    \end{tabular}%
  \label{VIF_compare_table}
  \vspace{-1.3em}
\end{table}

\subsection{Evaluation on Multi-Modal Fusion}
In this section, we compare our TC-MoA with ten task-specific image fusion methods and four general image fusion methods on the LLVIP dataset.

\begin{table}[t]\scriptsize
  \centering
\setlength{\tabcolsep}{2.7pt}

\caption{Quantitative results of the MEF task in MEFB \cite{MEF_Benchmark} . }
\vspace{-1.3em}
     \begin{tabular}{@{}lcccccccc@{}}
    \toprule
    Method & Psnr & $ Q_{c} $ & $ Q_{p} $ & $ Q_{g} $ & $ Q_{s} $ & MEF-SSIM & FMI & $ Q_{w} $ \\
    \midrule
    Deepfuse \cite{Deepfuse} & 57.104 & 0.426 & 0.352 & 0.325 & 0.641 & 0.897 & 0.873 & 0.548 \\
    MEF-GAN \cite{MEF-GAN}& 56.947 & 0.309 & 0.124 & 0.246 & 0.487 & 0.772 & 0.846 & 0.300 \\
    MEFNet \cite{MEF-Net} & 56.594 & {\color[HTML]{FF0000} \textbf{0.656}} & {\color[HTML]{0000FF} \textbf{0.595}} & {\color[HTML]{FF0000} \textbf{0.565}} & {\color[HTML]{FF0000} \textbf{0.838}} & 0.914 & {\color[HTML]{FF0000} \textbf{0.890}} & {\color[HTML]{FF0000} \textbf{0.866}} \\
    \midrule
    IFCNN \cite{IFCNN} & {\color[HTML]{0000FF} \textbf{57.195}} & 0.553 & 0.562 & 0.478 & 0.720 & {\color[HTML]{0000FF} \textbf{0.943}} & 0.882 & 0.834 \\
    FusionDN \cite{FusionDN} & 56.977 & 0.500 & 0.504 & 0.434 & 0.672 & 0.924 & 0.877 & 0.776 \\
    PMGI \cite{PMGI} & 57.117 & 0.489 & 0.525 & 0.442 & 0.666 & 0.936 & 0.885 & 0.804 \\
    U2Fusion \cite{U2Fusion} & 57.055 & 0.457 & 0.505 & 0.415 & 0.585 & 0.930 & 0.882 & 0.787 \\
    DeFusion \cite{DeFusion} & 57.131 & 0.539 & 0.378 & 0.376 & 0.751 & 0.902 & 0.877 & 0.733 \\
    TC-MoA & {\color[HTML]{FF0000} {\ul \textbf{57.213}}} & {\color[HTML]{0000FF} {\ul \textbf{0.578}}} & {\color[HTML]{FF0000} {\ul \textbf{0.598}}} & {\color[HTML]{0000FF} {\ul \textbf{0.528}}} & {\color[HTML]{0000FF} {\ul \textbf{0.767}}} & {\color[HTML]{FF0000} {\ul \textbf{0.964}}} & {\color[HTML]{0000FF} {\ul \textbf{0.888}}} & {\color[HTML]{0000FF} {\ul \textbf{0.845}}}
    \\
     \bottomrule
     \end{tabular}%
     \vspace{-2em}
  \label{MEF_compare_table}%
\end{table}%

\noindent\textbf{Quantitative Comparisons}. 
We evaluate the fusion performance on 8 quantitative metrics, as shown in Table~\ref{VIF_compare_table}. Our method outperforms all the general image fusion methods, demonstrating our superior compatibility across multiple fusion tasks.
For most task-specific image fusion methods, TC-MoA also performs remarkable improvements, although these methods deploy complex task-specific designs. 
Specifically, our method achieves a significant advantage in VIF, $Q_{cv}$, and EN metrics correlated to human perception or information theory. This indicates that our fused image conforms more to human visual perception and contains more information from the source images. The results on MS-SSIM, SD, and $Q_{w}$  metrics also explain that our fusion results possess a sufficient amount of structural information and gradient information.

\noindent\textbf{Qualitative Comparisons.} 
As shown in \cref{Vis_VIF}, our model outperforms the competing methods in terms of visual quality. For instance, the contours of the tree in dark areas are well depicted and the background texture is clearer. It is worth noting that our model has the capability to directly generate color images, or we can also follow other approaches to conduct on gray images and colorize the results to obtain TC-MoA$^{YUV}$. 
Compared with the competing methods, our model exhibits higher contrast between foreground and background, and more saturated colors. This comparison demonstrates our model is effective in generating fused images with superior perceptual quality.


%

\subsection{Evaluation on Multi-Exposure Fusion}

In this section, we compare TC-MoA with three task-specific MEF methods \ie Deepfuse\cite{Deepfuse}, MEF-GAN\cite{MEF-GAN} and MEF-Net \cite{MEF-Net}, and five general image fusion methods \ie IFCNN \cite{IFCNN}, FusionDN \cite{FusionDN}, PMGI \cite{PMGI}, U2Fusion \cite{U2Fusion}, and DeFusion \cite{DeFusion}.


\noindent\textbf{Quantitative Comparisons.} 
We employ 8 quantitative metrics to evaluate our model and the competing methods on MEFB, as presented in Table~\ref{MEF_compare_table}.  Our model achieves the SoTA performance in the general image fusion methods and achieves competitive results in task-specific methods. For example, our model significantly improves the MEF-SSIM scores due to the compatibility of TC-MoA with diverse tasks, enabling task-specific optimization while reducing task conflicts. The MEF-SSIM focuses on the structure and contrast distortion of images, and $Q_{p}$  measures the phase congruency of the source and fused images. The highest MEF-SSIM, PSNR, and $Q_{p}$ values indicate our effectiveness in the preservation of structural information, detailed features, and image quality in the fused images.

\begin{figure}[t]
  \centering
    \includegraphics[width=1\linewidth]{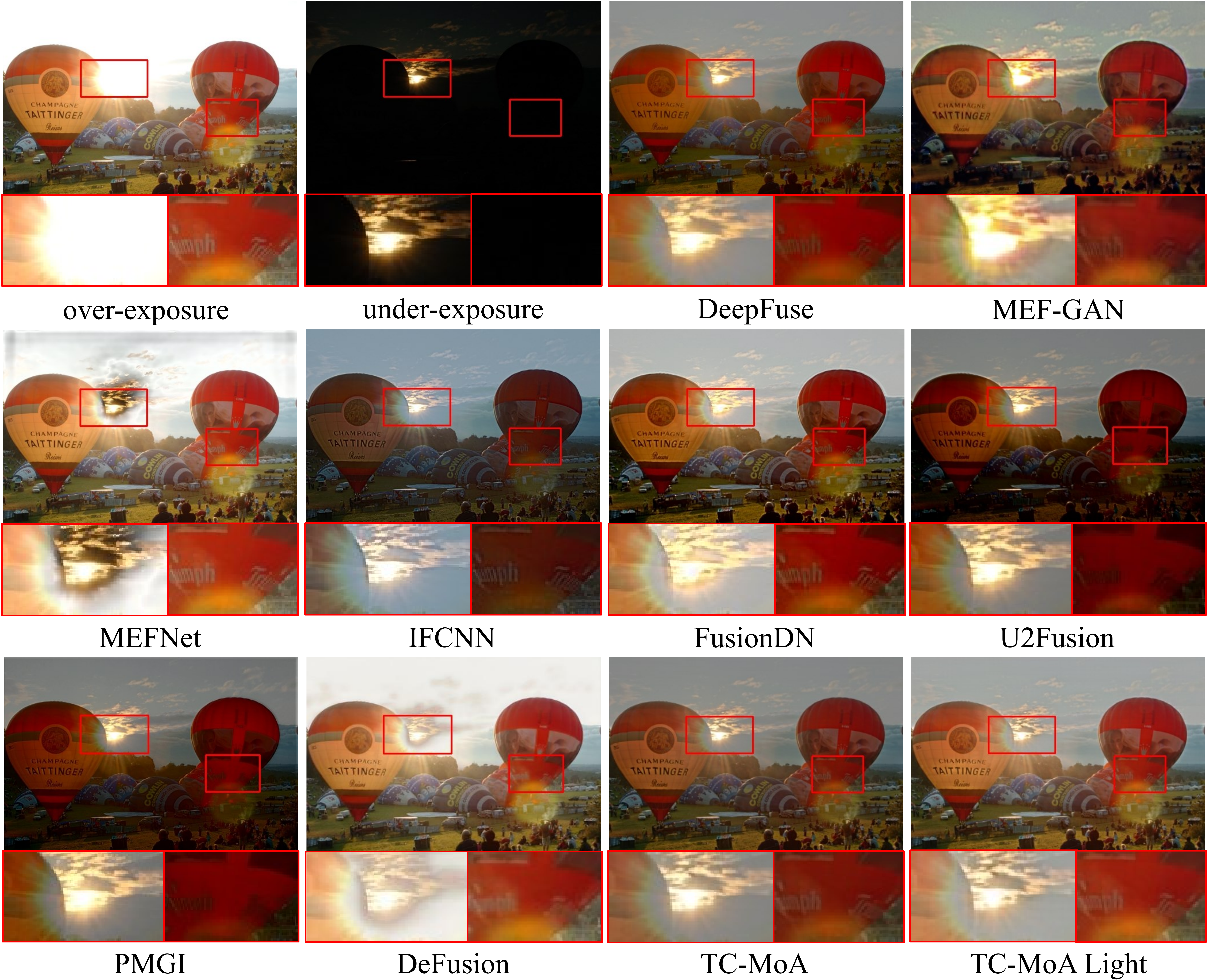}
    \vspace{-2em}
    \caption{Qualitative comparisons in MEF task.}
\vspace{-1.6em}
    \label{Vis_MEF}
\end{figure}

\noindent\textbf{Qualitative Comparisons.} As shown in \cref{Vis_MEF}, our method preserves the most texture details and color in both high and low illumination regions. Specifically, in the high illumination region, our model effectively retains more structural information of the clouds around the sun.  In the low illumination region, the colors of the fonts on the hot air balloon are completely confused in the PMGI and U2Fusion methods. As a comparison, our method maintains detailed information while keeping more accurate color information. In fact, there is no objective standard for the brightness of the fused image for multi-exposure images. To solve this problem, our method makes the fusion of images controllable by modulating the prompt. The image of “TC-MoA Light" shows the result of this modulation. Detailed information about the modulation can be found in \cref{PromptControllability}.

 
   

\begin{figure}[t]
  \centering
    \includegraphics[width=1\linewidth]{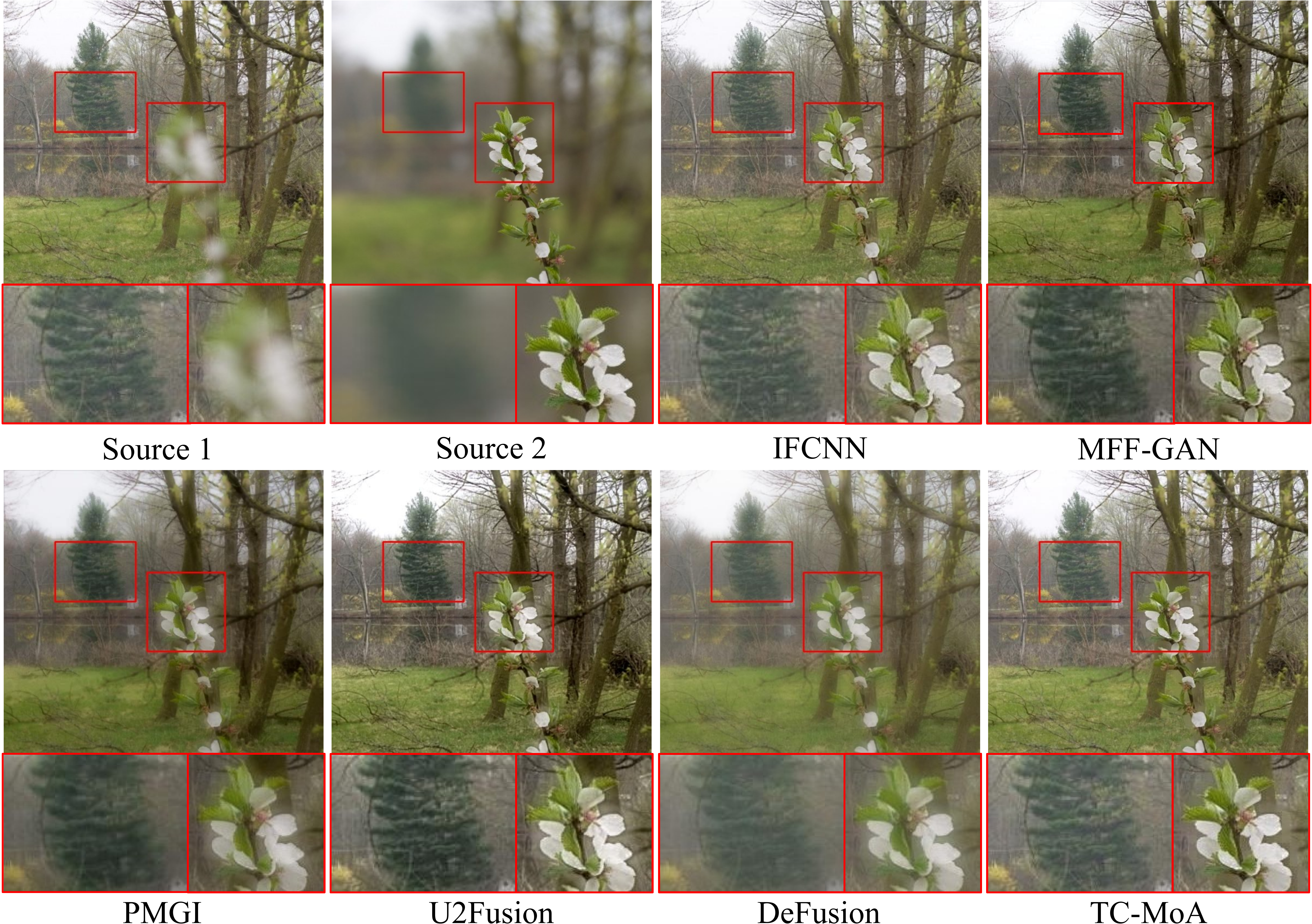}
    \vspace{-1.3em}
    \caption{Qualitative comparisons in MFF task.}
    \vspace{-1.0em}
    \label{Vis_MFF}
\end{figure}

\begin{table}[th]\scriptsize
    \centering
  \setlength{\tabcolsep}{2.55pt}

  \caption{Quantitative results of the MFF task.}
   \vspace{-1.3em}
     \begin{tabular}{@{}lcccccccc@{}}
    \toprule
     Method &    NMI & $ Q_{ncie} $ & MI & $ Q_{cb} $ & \textbf{$ Q_{cv} $↓} & MS-SSIM & FMI & $ Q_{w} $ \\
    \midrule
       IFCNN \cite{IFCNN}& {\color[HTML]{0000FF} \textbf{0.847}} & {\color[HTML]{0000FF} \textbf{0.825}} & {\color[HTML]{0000FF} \textbf{6.495}} & {\color[HTML]{0000FF} \textbf{0.691}} & {\color[HTML]{0000FF} \textbf{44.373}} & {\color[HTML]{FF0000} \textbf{0.991}} & {\color[HTML]{FF0000} \textbf{0.881}} & {\color[HTML]{FF0000} \textbf{0.912}} \\
       MFF-GAN \cite{MFI-WHU} & 0.728 & 0.820 & 5.689 & 0.609 & 72.460 & 0.977 & 0.876 & 0.877 \\
    \midrule
    FusionDN \cite{FusionDN}& 0.675 & 0.818 & 5.449 & 0.519 & 178.491 & 0.942 & 0.864 & 0.794 \\
    PMGI \cite{PMGI}& 0.702 & 0.819 & 5.522 & 0.542 & 140.957 & 0.923 & 0.866 & 0.575 \\
    U2Fusion \cite{U2Fusion}& 0.670 & 0.818 & 5.329 & 0.530 & 142.325 & 0.948 & 0.867 & 0.830 \\
    DeFusion \cite{DeFusion}& 0.768 & 0.821 & 5.838 & 0.625 & 83.195 & 0.960 & 0.870 & 0.685 \\
    TC-MoA & {\color[HTML]{FF0000} \textbf{0.875}} & {\color[HTML]{FF0000} \textbf{0.827}} & {\color[HTML]{FF0000} \textbf{6.695}} & {\color[HTML]{FF0000} \textbf{0.718}} & {\color[HTML]{FF0000} \textbf{36.512}} & {\color[HTML]{0000FF} \textbf{0.990}} & {\color[HTML]{0000FF} \textbf{0.881}} & {\color[HTML]{0000FF} \textbf{0.891}}
\\
    \bottomrule
    \end{tabular}%
  \label{MFF_compare_table}%
   \vspace{-2em}
\end{table}%

\subsection{Evaluation on Multi-Focus Fusion}



We compare TC-MoA with two task-specific MFF methods \ie  IFCNN \cite{IFCNN} (supervised on MFF) and MFF-GAN \cite{MFI-WHU}, and four general image fusion methods \ie FusionDN \cite{FusionDN}, PMGI \cite{PMGI}, U2Fusion \cite{U2Fusion}, and DeFusion \cite{DeFusion}. 

\noindent\textbf{Quantitative Comparisons.} 
We employ 8 quantitative metrics on the dataset provided in the \cite{MFF_DL_Survey},  as illustrated in Table~\ref{MFF_compare_table}. Our method shows competitive performance compared to most image fusion methods. 
For general image fusion, our model achieves superior performance across all metrics. Additionally, TC-MoA directly generates the fused image without predicting the decision map. Nevertheless, we still significantly outperform IFCNN in NMI, MI, Qcb, and Qcv metrics. This indicates that our fusion results, compared to supervised method, retain more source image details and demonstrate remarkable human visual perception.


\noindent\textbf{Qualitative Comparisons.} 
As depicted in \cref{Vis_MFF}, our fused image exhibits superior consistency in terms of texture and color, surpassing that of the other methods. U2Fusion exhibits color deviations in the far-focus region, while IFCNN distorts the near-focus image. Furthermore, these methods are unable to effectively remove the distortion around the flowers,  while we introduce an MFF-specific loss, which achieves the most visually appealing results for the flower.

\subsection{Ablation Study}
We conduct ablation studies to verify the effect of our TC-MoA 
and network architecture.
\noindent\textbf{TC-MoA.} 
To verify the effectiveness of TC-MoA, we remove it from our framework. As shown in the first and second rows of Table~\ref{Ablation_single_mulit_Lp}, 
the models trained by multi-task training outperform those trained for a single task, based on 5 commonly used metrics. Interestingly though, this rule reverses in the case of MEF, indicating that inter-task competition occurs for models trained directly with multiple tasks. Comparatively, the model incorporating the multiple adapters improves the performance across all metrics, suggesting that TC-MoA is dynamically compatible with different tasks.

\noindent\textbf{Adapter-based Fine-tuning.} 
We performed ablation experiments on the backbone and adapter methods, as shown in Table~\ref{Ablation_fine-tuning}. The fine-tuning method of “FrozenBackbone” represents freezing the entire backbone and only unfreezing the final linear layer. 
Compared to the approach of “FrozenBackbone”, our TC-MoA achieves significant fusion performance improvements by introducing a small number of learnable parameters. More importantly, our method outperforms the existing adapter-based visual fine-tuning approach AdaptFormer~\cite{AdaptFormer} on both base and large versions of pre-trained models, demonstrating the superiority of the TC-MoA structure in fusion tasks. 
In addition to this, 
Table~\ref{VIF_TNO_Exp} shows our base version achieves competitive performance on the VIF task, which can be further enhanced by the large version. 

\begin{table}[t]\scriptsize
    \centering
  \setlength{\tabcolsep}{4.8pt}
  \caption{Ablation studies on TC-MoA. S and M represent single task training and multi-task training, while SA and MA denote single adapter and multiple adapters respectively. }
  \vspace{-1.3 em}
\begin{tabular}{@{}c|cccc|ccccc}
\toprule
Task & S & M & SA & MA  & $Q_{abf}$ & $Q_{p}$ & FMI & $Q_{c}$ & SSIM \\ \midrule
 & \checkmark &  & \checkmark &   & 0.5984 & 0.4073 & 0.8857 & 0.6335 & 0.4540 \\
 &  & \checkmark & \checkmark &   & 0.5997 & 0.4116 & {\color[HTML]{FF0000} \textbf{0.8862}} & 0.6357 & 0.4544 \\
 \multirow{-3}{*}{VIF} & \cellcolor[HTML]{EFEFEF} & \cellcolor[HTML]{EFEFEF}\checkmark & \cellcolor[HTML]{EFEFEF} & \cellcolor[HTML]{EFEFEF}\checkmark  & \cellcolor[HTML]{EFEFEF}{\color[HTML]{FF0000} \textbf{0.6007}} & \cellcolor[HTML]{EFEFEF}{\color[HTML]{FF0000} \textbf{0.4119}} & \cellcolor[HTML]{EFEFEF}0.8862 & \cellcolor[HTML]{EFEFEF}{\color[HTML]{FF0000} \textbf{0.6365}} & \cellcolor[HTML]{EFEFEF}{\color[HTML]{FF0000} \textbf{0.4550}} \\
 \hline
 & \checkmark &  & \checkmark &   & 0.6385 & 0.5916 & 0.8875 & 0.5754 & 0.4115 \\
 &  & \checkmark & \checkmark &   & 0.6362 & 0.5899 & 0.8875 & 0.5765 & 0.4106 \\
 \multirow{-3}{*}{MEF} & \cellcolor[HTML]{EFEFEF} & \cellcolor[HTML]{EFEFEF}\checkmark & \cellcolor[HTML]{EFEFEF} & \cellcolor[HTML]{EFEFEF}\checkmark  & \cellcolor[HTML]{EFEFEF}{\color[HTML]{FF0000} \textbf{0.6449}} & \cellcolor[HTML]{EFEFEF}{\color[HTML]{FF0000} \textbf{0.5980}} & \cellcolor[HTML]{EFEFEF}{\color[HTML]{FF0000} \textbf{0.8883}} & \cellcolor[HTML]{EFEFEF}{\color[HTML]{FF0000} \textbf{0.5776}} & \cellcolor[HTML]{EFEFEF}{\color[HTML]{FF0000} \textbf{0.4116}} \\
 \hline
 & \checkmark &  & \checkmark &   & 0.6517 & 0.6733 & 0.8797 & 0.7702 & 0.6774 \\
 &  & \checkmark & \checkmark &   & 0.6562 & 0.6799 & 0.8806 & 0.7742 & 0.6792 \\
 \multirow{-3}{*}{MFF} & \cellcolor[HTML]{EFEFEF} & \cellcolor[HTML]{EFEFEF}\checkmark & \cellcolor[HTML]{EFEFEF} & \cellcolor[HTML]{EFEFEF}\checkmark  & \cellcolor[HTML]{EFEFEF}{\color[HTML]{FF0000} \textbf{0.6568}} & \cellcolor[HTML]{EFEFEF}{\color[HTML]{FF0000} \textbf{0.6811}} & \cellcolor[HTML]{EFEFEF}{\color[HTML]{FF0000} \textbf{0.8808}} & \cellcolor[HTML]{EFEFEF}{\color[HTML]{FF0000} \textbf{0.7755}} & \cellcolor[HTML]{EFEFEF}{\color[HTML]{FF0000} \textbf{0.6794}} \\
\bottomrule
\end{tabular}
\label{Ablation_single_mulit_Lp}%
\vspace{-1.3em}
\end{table}

\subsection{Analysis and Discussion} \label{PromptControllability}

\noindent\textbf{Efficiency}. 
We have observed the speed issues brought about by the massive parameters of the pre-trained models. To address this, we have optimized the ViT architecture by the shifted windows, which will be detailed in \cref{NetworkDetail}. After optimization, for multi-task with arbitrary-size inputs, Table~\ref{Efficiency} shows the base and large versions of the model have accelerated by 178\% and 167\%,  resulting in acceptable FPS (Frames Per Second) compared with other methods. The “Frozen” refers to the vanilla ViT architecture. 
\begin{equation}
\begin{split}
  prompt_{x}^{'}  = \mu +\alpha (prompt_{x}-\mu )+\beta  \\
  prompt_{y}^{'}  = \mu +\alpha (prompt_{y}-\mu )-\beta 
  \label{prompt_scale}
  \end{split}
\end{equation}

\begin{figure*}[t]
  \centering
 
    \includegraphics[width=1\linewidth]
    {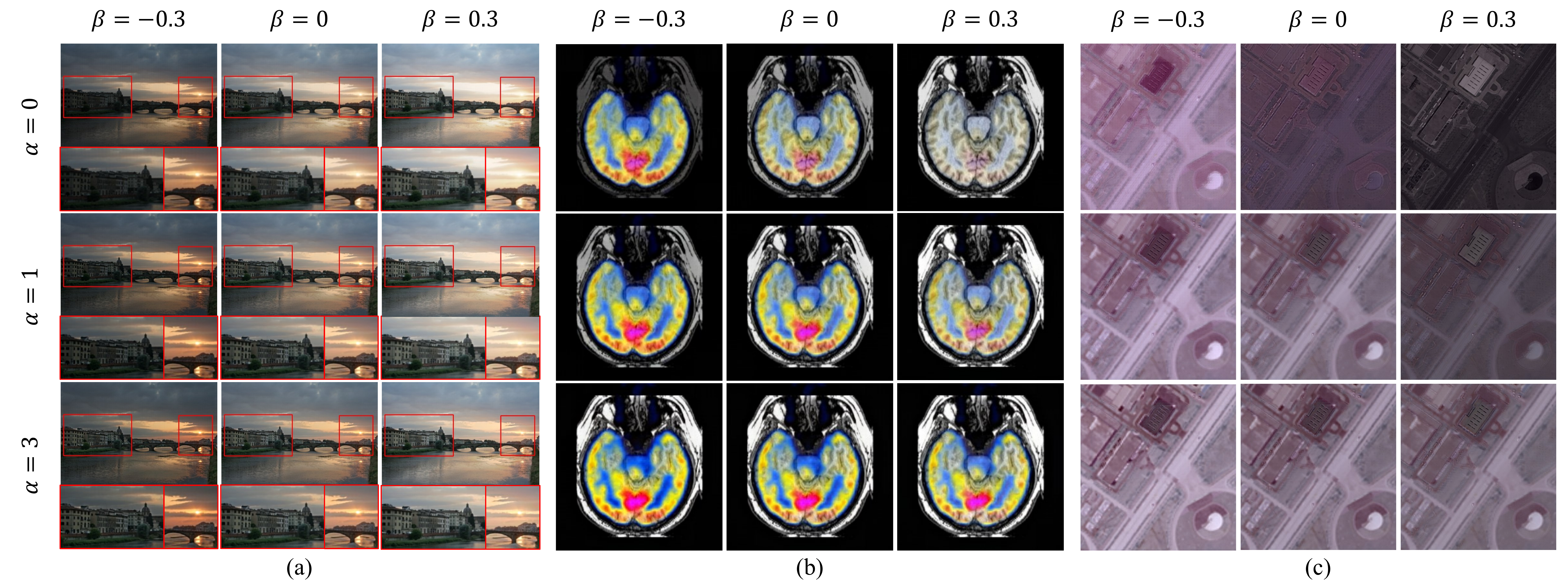}
    \vspace{-2em}
    \caption{ Visualisation of controllability and generalization of our method.  }
    \label{Vis_Scale}
    \vspace{-1.3em}
\end{figure*}

\begin{figure}[t]
  \centering
    \includegraphics[width=1\linewidth]{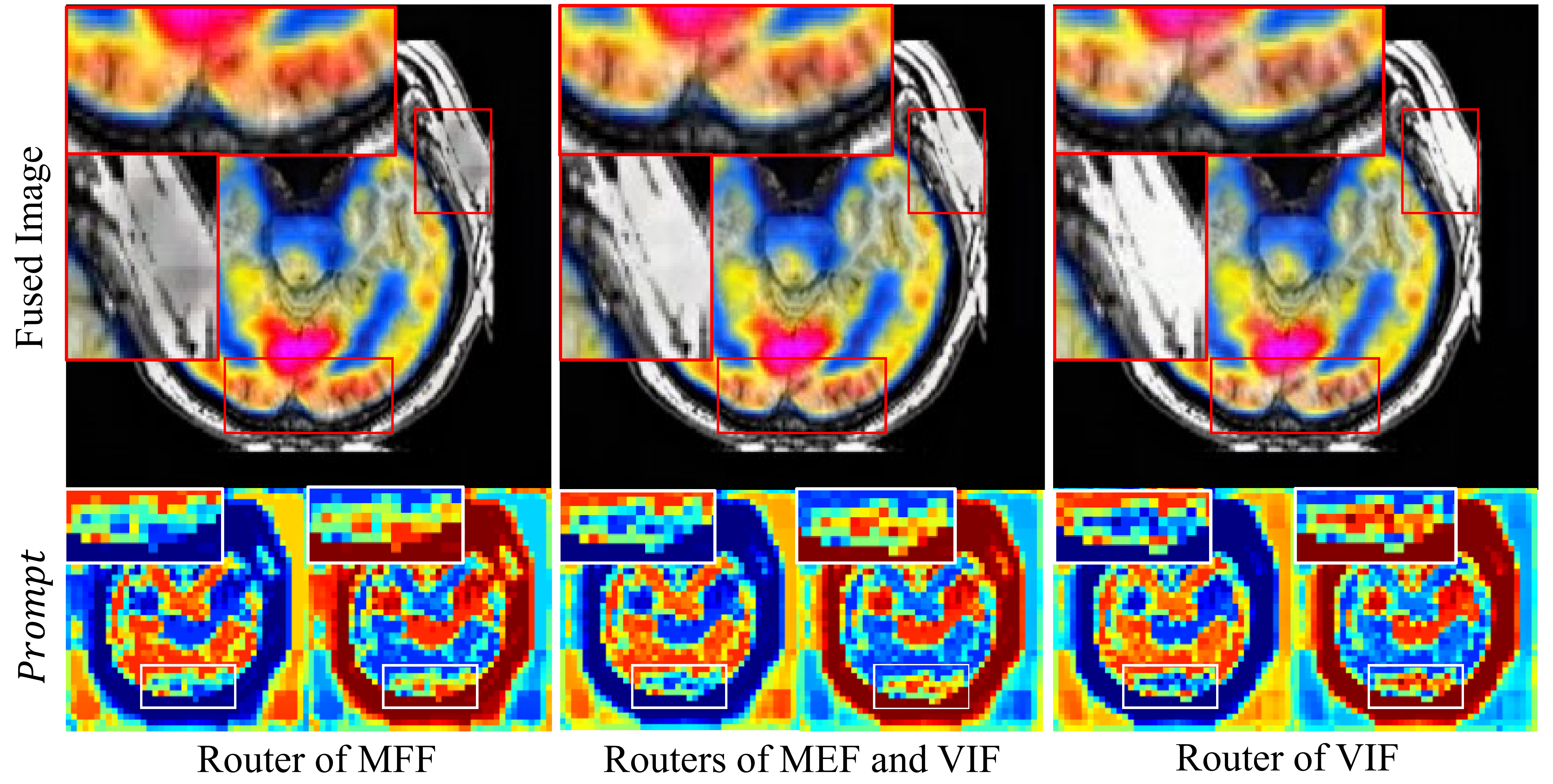}
    \vspace{-1.6em}
    \caption{Visualisation of different routers on medical images. }
    \label{Vis_diff_router}
    \vspace{-1.1em}
\end{figure}

\begin{table}[t]\scriptsize
    \centering
  \setlength{\tabcolsep}{3pt}
  \caption{Ablation studies on fine-tuning. $\mathbb{B} $ and $\mathbb{L} $ represent the use of ViT-base and ViT-large structures respectively.  }
  \vspace{-1.3em}
\resizebox{1.0\columnwidth}{!}{
\begin{tabular}{@{}l|ccc|ccc|ccc@{}}
\toprule
\multicolumn{1}{c|}{} & \multicolumn{3}{c|}{VIF} & \multicolumn{3}{c|}{MEF} & \multicolumn{3}{c}{MFF} \\ \cline{2-10} 
\multicolumn{1}{c|}{\multirow{-2}{*}{Architecture}} & \cellcolor[HTML]{EFEFEF}$Q_{abf}$ & \cellcolor[HTML]{EFEFEF}$Q_{p}$ & \cellcolor[HTML]{EFEFEF}SSIM & \cellcolor[HTML]{EFEFEF}$Q_{abf}$ & \cellcolor[HTML]{EFEFEF}$Q_{p}$ & \cellcolor[HTML]{EFEFEF}SSIM & \cellcolor[HTML]{EFEFEF}$Q_{abf}$ & \cellcolor[HTML]{EFEFEF}$Q_{p}$ & \cellcolor[HTML]{EFEFEF}SSIM \\ \midrule
FrozenBackbone$_\mathbb{B} $ & 0.175 & 0.089 & 0.275 & 0.250 & 0.139 & 0.227 & 0.295 & 0.189 & 0.466 \\
AdaptFormer$_\mathbb{B} $ & 0.531 & 0.336 & 0.427 & 0.574 & 0.504 & 0.377 & 0.579 & 0.561 & 0.651 \\
TC-MoA$_\mathbb{B} $ & {\color[HTML]{0000FF} \textbf{0.576}} & {\color[HTML]{0000FF} \textbf{0.396}} & {\color[HTML]{0000FF} \textbf{0.450}} & 0.604 & 0.540 & 0.394 & 0.612 & 0.619 & 0.660 \\ \midrule
FrozenBackbone$_\mathbb{L} $ & 0.330 & 0.238 & 0.363 & 0.452 & 0.408 & 0.333 & 0.450 & 0.428 & 0.596 \\
AdaptFormer$_\mathbb{L} $   & 0.576 & 0.392 & 0.446 & {\color[HTML]{0000FF} \textbf{0.616}} & {\color[HTML]{0000FF} \textbf{0.563}} & {\color[HTML]{0000FF} \textbf{0.396}} & {\color[HTML]{0000FF} \textbf{0.640}} & {\color[HTML]{0000FF} \textbf{0.650}} & {\color[HTML]{0000FF} \textbf{0.671}} \\
TC-MoA$_\mathbb{L} $  & {\color[HTML]{FF0000} \textbf{0.601}} & {\color[HTML]{FF0000} \textbf{0.412}} & {\color[HTML]{FF0000} \textbf{0.455}} & {\color[HTML]{FF0000} \textbf{0.645}} & {\color[HTML]{FF0000} \textbf{0.598}} & {\color[HTML]{FF0000} \textbf{0.412}} & {\color[HTML]{FF0000} \textbf{0.657}} & {\color[HTML]{FF0000} \textbf{0.681}} & {\color[HTML]{FF0000} \textbf{0.679}} \\
\bottomrule
\end{tabular}
}
\label{Ablation_fine-tuning}%
\vspace{-1.3em}
\end{table}

\begin{table}[t]\scriptsize
    \centering
  \setlength{\tabcolsep}{3pt}
  \caption{Efficiency and scalability comparisons of approaches. }
  \vspace{-1.3em}
\resizebox{1.0\columnwidth}{!}{
\begin{tabular}{@{}l|ccccccc@{}}
\toprule
& U2F & DDFM & CDD & Frozen$_\mathbb{B}$ &  TC-MoA$_\mathbb{B} $ & Frozen$_\mathbb{L} $ & TC-MoA$_\mathbb{L} $ \\ 
\midrule
Total Params (M) & 1.32 & - & 1.78 & 111.65 & 115.40 & 339.12 & 348.70 \\
Trainable Params (M) & 1.32 & - & 1.78 & - & 3.87 & - & 9.58 \\ 
FPS (VIF 640$\times$512)
 & 4.72 & 0.01 & 2.52 & 1.17 & 3.33 & 0.60 & 1.60 \\
FPS (MEF arbitrary-size)
 & 2.21 & - & - & 1.37 & 3.58 & 0.71 & 1.86 \\
FPS (MFF arbitrary-size)
 & 1.86 & - & - & 1.39 & 3.98 & 0.71 & 1.95 \\
\bottomrule
\end{tabular}
}
\label{Efficiency}%
\vspace{-2em}
\end{table}

 

\noindent\textbf{Prompt Controllability}. 
Employing the formula noted in \cref{prompt_scale}, we manipulate both the scale and shift of the prompt provided to the trained model, where $ \mu=0.5 $ represents the mean, while $ \alpha $ and $ \beta $ denote the scaling and shifting factors, respectively. We conduct the manipulation on some samples and present an example in \cref{Vis_Scale} (a).

Obviously, as the value of $\beta$ increases, the brightness and texture of the fused image increasingly resemble the source image $ X $, and inversely, the source image $ Y $.  Thus, $ \beta $ can be considered as a shifting of the dominant intensity bias. Moreover, a higher value of $ \alpha $ tends to favor one source image over the other in the image patches, which can be considered as a scaling of the dominant intensity deviation.
For example, outputs with $\beta$ = -0.3 are more similar to the under-exposure image globally. As $ \alpha $ increases from 0 to 3, the building areas of the fused image gradually become brighter, but the brightness of the clouds remains basically unchanged. This suggests that changing the scale factor of the prompt motivates the model to select different regions. Overall, this experiment demonstrates that the prompt is controllable and can be linearly transformed to obtain fuse images in different degrees.
\noindent\textbf{Generalizability.} 
Our method is highly generalizable to tasks with similar fusion rules by zero-shot fusion. Specifically, we take samples from medical image fusion dataset~\cite{AANLIB} and pan-sharpening dataset generated by the Quickbird satellite, which are fused based on VIF's paradigm. We obtain a base fused image with $ \alpha=1 $ and $ \beta=0 $, as shown in \cref{Vis_Scale} (b) and (c). However, without any prior knowledge of these unknown tasks, the output fused image is an under-performing image, missing many high and low-frequency information. 
Interestingly, by manipulating the prompt, we can find a suitable fusion rule for these new tasks, obtaining reasonable fusion results.


\noindent\textbf{Routers Controllability.} 
As shown in \cref{Vis_diff_router}, the images generated by the MFF router are darker than VIF in areas without gradient guidance. The MFF selects the reference sources based on gradients, it tends to average the information from sources without gradient guidance. In contrast, the VIF router tends to preserve the maximum intensity information. Moreover, at the bottom of the image, the MFF router tends to preserve the gradient information depending on a single source, while the VIF router preserves the gradient information from both sources. Additionally, by manipulating the combination of different routers, our model copes with controlling the dynamic changing fusion degree, obtaining more controllable results.

\begin{figure}[t]
  \centering
    \includegraphics[width=1\linewidth]{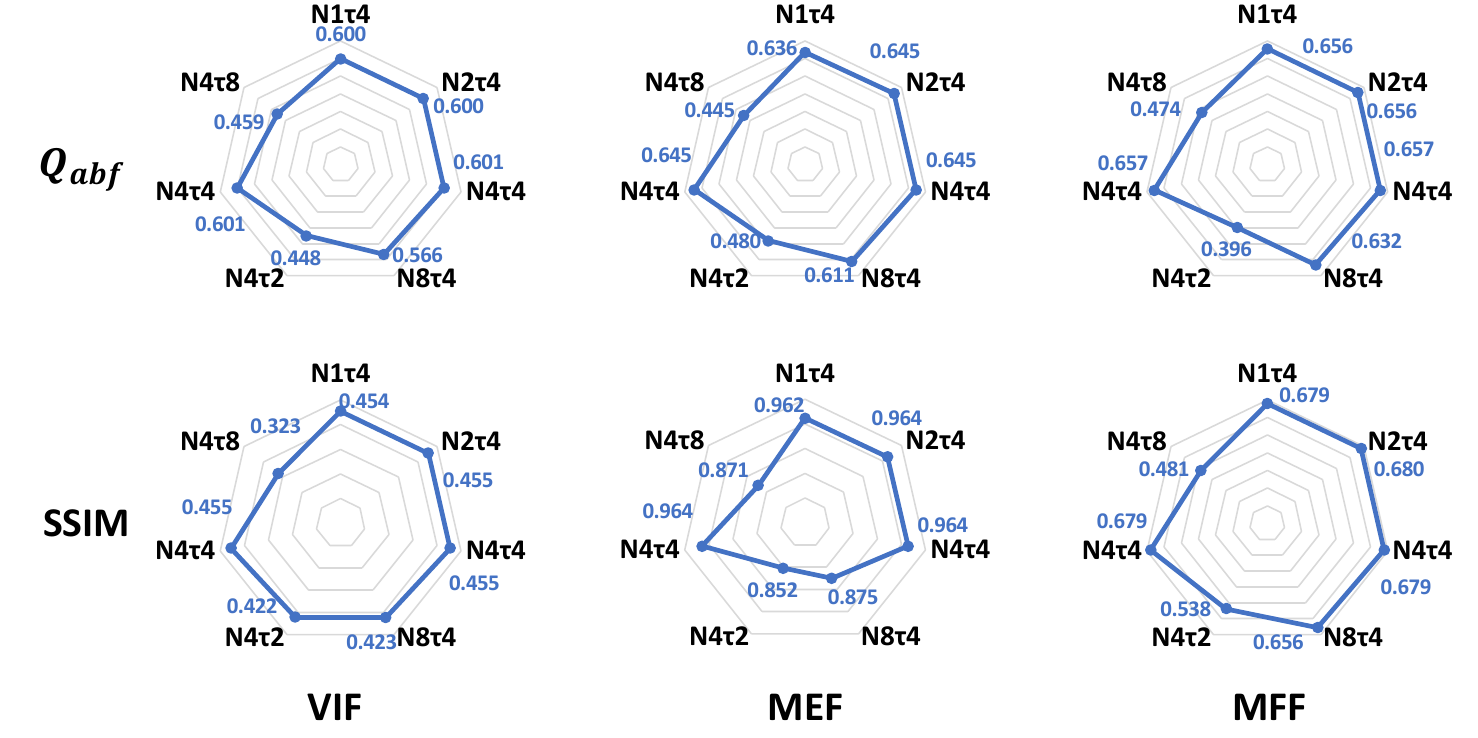}
    \vspace{-1.5em}
    \caption{Experiments on the hyperparameters.
    }
    \label{Vis_N_tau}
    \vspace{-1.1em}
\end{figure}

\noindent\textbf{Hyperparameters.} 
We conducted hyperparameters analysis on two aspects: the number of experts ($N$) and the interval between two TC-MoAs ($\tau$). We utilize the ViT-large as our backbone, which consists of 24 transformer blocks in the encoder and 8 in the decoder. If $\tau=4 $, our model contains 8 instances of TC-MoA. We employ two widely-used metrics $Q_{abf}$ and SSIM to inform the selection of hyperparameters. As presented in \cref{Vis_N_tau}, as $ N $ increases, the performance initially increases but eventually declines, with the peak performance detected at $ N=4 $. This suggests that increasing the number of experts does not necessarily enhance the model's performance. Similarly, when $ \tau=4 $, the model demonstrates the best performance across multi-task. Hence, our experiments are conducted under $ N=4 $ and $ \tau=4 $. 

%% file: arxiv/sec/5_Conclusion.tex
\section{Conclusion}

In this paper, we propose a task-customized mixture of adapters for general image fusion. With the help of TC-MoA, different fusion tasks customize their respective mixture of adapters to obtain prompt guidance for multiple-source image fusion in a unified model.
To ensure compatibility with different tasks while maintaining complementarity for multi-source images, we further propose mutual information regularization to constrain these adapters. 
Experimental results have shown that TC-MoA achieves advanced performance in general image fusion against the competing methods. In addition, TC-MoA demonstrates strong prompt controllability and router controllability to perform flexible manipulation on the fused images.


%% file: arxiv/sec/6_Acknowledgement.tex
\section{Acknowledgement}

This work was sponsored in part by the National Key R\&D Program of China 2022ZD0116500, in part by the National Natural Science Foundation of China (62222608, 62106171, U23B2049, 61925602), in part by the Haihe Lab of ITAI under Grant 22HHXCJC00002, in part by the Tianjin Natural Science Foundation under Grant 21JCYBJC00580, in part by Tianjin Natural Science Funds for Distinguished Young Scholar under Grant 23JCJQJC00270, and in part by the Key Laboratory of Big Data Intelligent Computing, Chongqing University of Posts and Telecommunications under Grant BDIC-2023-A-008.
This work was also sponsored by CAAI-CANN Open Fund, developed on OpenI Community.




%% file: arxiv/sec/X_suppl.tex
\clearpage
\setcounter{page}{1}
\maketitlesupplementary
\begin{abstract}
In supplementary material, we provide more details, discussions and experiments for the paper “Task-Customized Mixture of Adapters for General Image Fusion", encompassing the following:

(\uppercase\expandafter{\romannumeral1}) A comprehensive overview of the task-customized loss functions along with associated comparative experiments, as detailed in \cref{lossfunction}.

(\uppercase\expandafter{\romannumeral2}) More details about the network. This includes the detailed network structure and data flow of TC-MoA, the implementation method of shifted windows, the ablation studies about network design and the analysis of parameters, elaborated in \cref{NetworkDetail}. 

(\uppercase\expandafter{\romannumeral3}) Additional analysis and discussion concerning the properties exhibited by different tasks in the experiments and an exploration into the phenomenon of task-specific routing, explored in \cref{AdditionalAnalysis}. 

(\uppercase\expandafter{\romannumeral4}) More quantitative results on TNO dataset, as shown in \cref{Morequantitative}. And more qualitative results of fused images of multiple tasks, as presented in \cref{Morequalitative}. 

\end{abstract}


\section{Task-Customized Loss Function}\label{lossfunction}




Our network structure accommodates the unique requirements of varied tasks, with tailored unsupervised loss functions for each fusion task. 
In order to generate high-quality fused images, we impose constraints on the structural ($\mathcal{L} _{ssim}$), intensity ($\mathcal{L} _{Pixel}$), and gradient information ($\mathcal{L} _{Grad}$) of the fused images for different fusion tasks.


Specifically, for the VIF task, our primary goal is to retain the most distinct high-frequency and low-frequency information from the source images. With this regard, we introduce $\mathcal{L} _{MaxPixel}$ and $\mathcal{L} _{MaxGrad}$ loss functions in this task. By employing $\mathcal{L} _{MaxPixel}$ \cite{SeAFusion} loss function, the fused images have more comprehensive shapes of objects in the dark areas and better color saturation.  To maintain gradient information, we ensure the gradient's sign (direction) remains unchanged in all related loss functions to avoid any unintended gradient confusion.
\begin{equation}
  \mathcal{L} _{V} =\mathcal{L} _{aux}+ \mathcal{L} _{ssim} + \mathcal{L} _{MaxPixel}+\mathcal{L} _{MaxGrad},
  \label{eq:Lv_ref}
\end{equation}
\begin{equation}
    \begin{split}
    \mathcal{L}_{ssim} = \lambda_{1} (1-SSIM(I_{Fusion},X)) \\
    +\lambda_{2}(1-SSIM(I_{Fusion},Y)),
  \label{eq:Lssim}
  \end{split}
\end{equation}
\begin{equation}
  \mathcal{L}_{MaxPixel} = \frac{1}{HW}\left \| I_{Fusion}-max(X,Y) \right \|_1,
  \label{eq:LMaxPixel}
\end{equation}
\begin{equation}
  \mathcal{L}_{MaxGrad} = \frac{1}{HW}\left \|   \nabla I_{Fusion} - absmax(\nabla X,\nabla Y) \right \|_1,
  \label{eq:LMaxGrad}
\end{equation}
\noindent where $ \mathcal{L} _{aux} $ is the auxiliary loss to avoid  unbalanced learning of adapters. 
$ \mathcal{L} _{ssim} $  represents the loss function based on the structural similarity (SSIM) metric, where $ \lambda_{1}$ and $\lambda_{2} $ are set to $0.5$. 
The $ mean(\cdot) $, $ max(\cdot) $, and $ absmax(\cdot) $ represent functions that compute the element-wise average, take the maximum selection, and get the maximum absolute value, respectively.
The Sobel operator is denoted as $\nabla$. For gradient-related loss functions like $ \mathcal{L} _{MaxGrad}$, we retain the sign of the gradient values.

\begin{table}[t]\scriptsize
\setlength{\tabcolsep}{5.3pt}
\centering
\begin{tabular}{@{}c|c|cccccc@{}}
\toprule
Task & \begin{tabular}[c]{@{}c@{}}Task-Customized\\ Loss\end{tabular} & $Q_{abf}$ & $\mathcal{VIF}$ & SSIM & $Q_p$ & $Q_c$ & $Q_{cb}$ \\ \midrule
 &  & 0.390 & 0.553 & 0.400 & 0.310 & 0.553 & 0.413 \\
\multirow{-2}{*}{VIF} & \checkmark & {\color[HTML]{FF0000} \textbf{0.601}} & {\color[HTML]{FF0000} \textbf{0.726}} & {\color[HTML]{FF0000} \textbf{0.455}} & {\color[HTML]{FF0000} \textbf{0.412}} & {\color[HTML]{FF0000} \textbf{0.637}} & {\color[HTML]{FF0000} \textbf{0.494}} \\ \midrule
 &  & 0.521 & 0.601 & 0.939 & 0.555 & 0.541 & 0.396 \\
\multirow{-2}{*}{MEF} & \checkmark & {\color[HTML]{FF0000} \textbf{0.645}} & {\color[HTML]{FF0000} \textbf{0.661}} & {\color[HTML]{FF0000} \textbf{0.964}} & {\color[HTML]{FF0000} \textbf{0.598}} & {\color[HTML]{FF0000} \textbf{0.578}} & {\color[HTML]{FF0000} \textbf{0.431}} \\ \midrule
 &  & 0.473 & 0.761 & 0.674 & 0.482 & 0.696 & 0.638 \\
\multirow{-2}{*}{MFF} & \checkmark & {\color[HTML]{FF0000} \textbf{0.657}} & {\color[HTML]{FF0000} \textbf{0.898}} & {\color[HTML]{FF0000} \textbf{0.679}} & {\color[HTML]{FF0000} \textbf{0.681}} & {\color[HTML]{FF0000} \textbf{0.775}} & {\color[HTML]{FF0000} \textbf{0.718}} \\ 
\bottomrule
\end{tabular}
\vspace{-1em}
\caption{Quantitative results of the task-customized loss functions. The SSIM metric in MEF task is replaced by MEF-SSIM metric.}
\vspace{-1.8em}
\label{tab:SameLoss_table}
\end{table}

For the MEF task, the fused images should maintain average luminance levels while retaining all gradient information. This strategy drives us to design $\mathcal{L} _{AvgPixel}$ and ${L} _{MaxGrad}$ loss functions for the MEF task. Additionally, we adopt $\mathcal{L} _{mefssim}$  which is specifically designed for the MEF task, instead of traditional $\mathcal{L} _{ssim}$.
\begin{equation}
  \mathcal{L} _{E} =\mathcal{L} _{aux}+ \mathcal{L} _{mefssim} + \mathcal{L} _{AvgPixel}+\mathcal{L} _{MaxGrad}
  \label{eq:LE}
\end{equation}
\begin{equation}
  \mathcal{L}_{AvgPixel} = \frac{1}{HW}\left \| I_{Fusion}-mean(X,Y) \right \|_1
  \label{eq:LAvgPixel}
\end{equation}

For the MFF task, each patch of the fused images tends to depend on one specific source image with the maximum gradient. This prevents the objects' edges in defocused areas from being preserved, thereby affecting the quality of the fused images. In practicality, we select only one source patch to calculate the loss function for each patch in fused images. Therefore, the $\mathcal{L} _{MaskPixel}$ and $\mathcal{L} _{MaskGrad}$ loss functions for MFF are designed as, 
\begin{equation}
  \mathcal{L} _{F} =\mathcal{L} _{aux}+ \mathcal{L} _{ssim} + 
  \mathcal{L} _{MaskPixel}+\mathcal{L} _{MaskGrad}
  \label{eq:LF}
\end{equation}
\begin{equation}
  \mathcal{L} _{MaskPixel} = 
  \sum_{i=1}^{2} M_i \circ  \left \| I_{Fusion}-I_i \right \|_1
  \label{eq:LMaskPixel}
\end{equation}
\begin{equation}
  \mathcal{L} _{MaskGrad} = 
  \sum_{i=1}^{2} M_i \circ \left \|   \nabla I_{Fusion} -  \nabla I_i \right \|_1
  \label{eq:LMaskGrad}
\end{equation}

\begin{figure*}[t]
  \centering
    \includegraphics[width=0.8\linewidth]{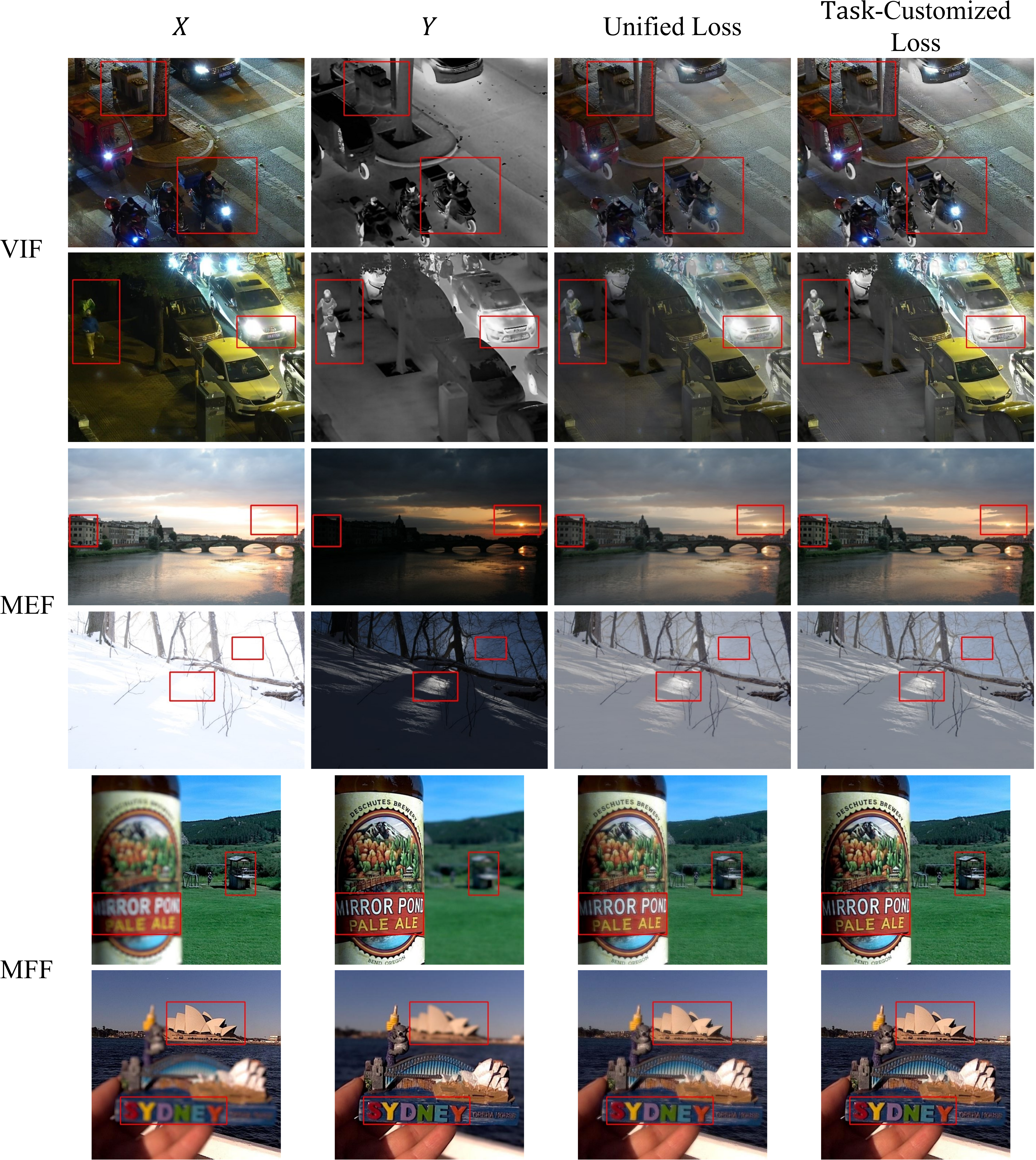}
     \vspace{-0.5em}
    \caption{Qualitative comparisons on task-customized loss functions. }
    \label{SameLoss_Img}
    \vspace{-1.3em}
\end{figure*}
\begin{figure*}[t]
  \centering
    \includegraphics[width=1\linewidth]{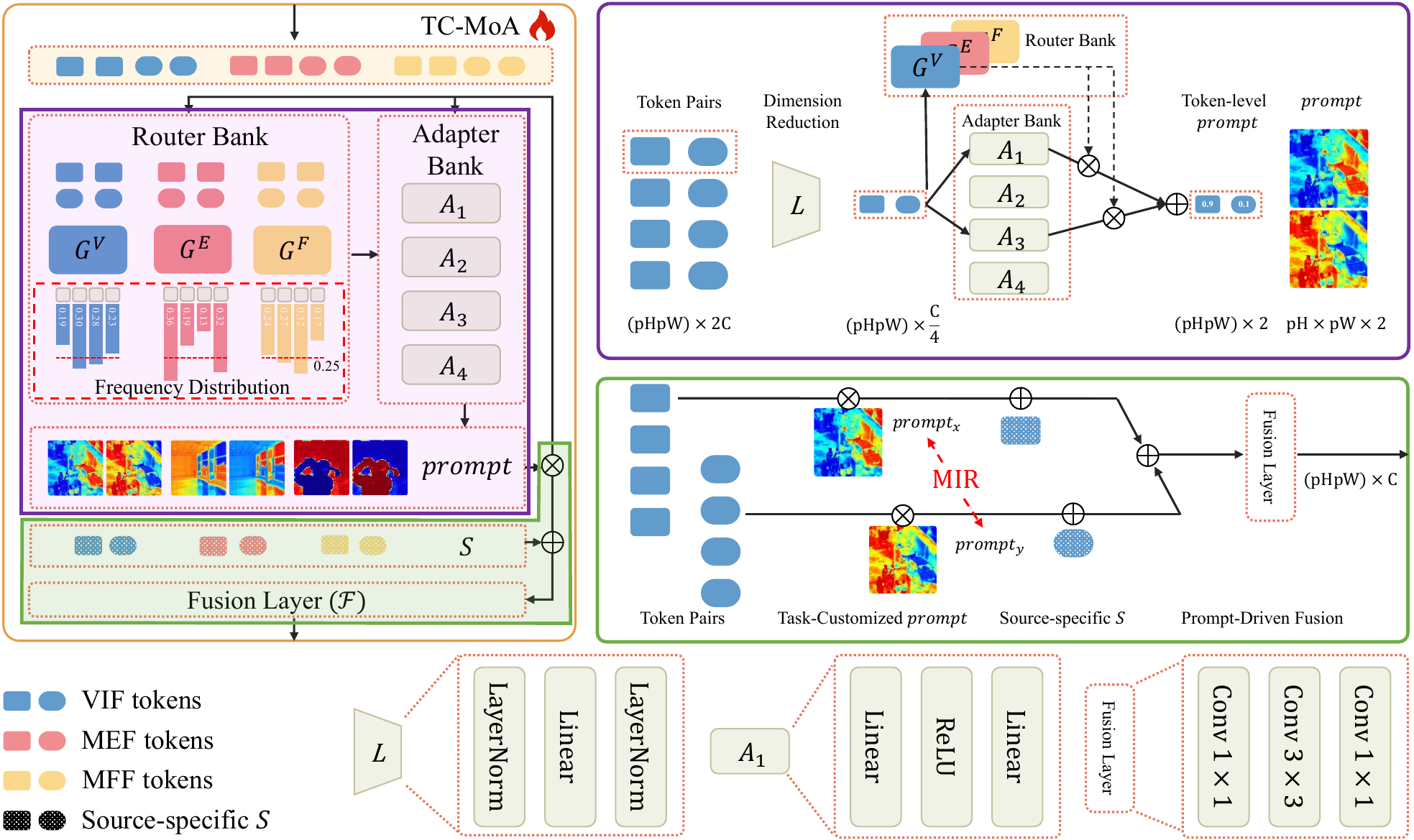}
    \vspace{-1.5em}
    \caption{Detail architecture for TC-MoA. The purple box represents the process of generating task-customized prompt, while the green box illustrates the prompt-driven fusion process. The MIR stands for mutual information regularization constraint. 
    The routing results of all samples for a specific task by a task-specific router constitute the frequency distribution of the mixture of adapters.
    }
    \vspace{-1.3em}
    \label{DetailNetWork}
\end{figure*}

where $ M $  represents the mask map $ M\in \mathbb{R}^{pH \times pW \times 1}$, and $ i $ denotes the index of image in source images tuple $ (X,Y) $. If the patch's maximum gradient in current image exceeds the other image, we allot the mask map value of the location of this patch as 1. Conversely, it is set to 0.


\noindent\textbf{Qualitative and Quantitative Comparisons.} 
We compare the model trained with task-customized loss functions and that trained with unified loss function. 
The unified loss function follows the PMGI combined with $\mathcal{L}_{ssim}$. The quantitative results are shown in Table~\ref{tab:SameLoss_table}, and the qualitative results are reported in~\cref{SameLoss_Img}.
The quantitative results show that the fused images obtained by our task-customized loss functions are rich in high frequency and structural details, conforming to human perception across all tasks. The qualitative results show that our fused images display superior contrast, color saturation, and textural detail when compared to images produced by unified loss function. It is worth noting that previous MFF methods often require post-processing on the fused images or features to obtain clear images, but our model can directly reconstruct the fused images with task-customized loss functions.


\begin{figure}[t]
  \centering
    \includegraphics[width=1\linewidth]{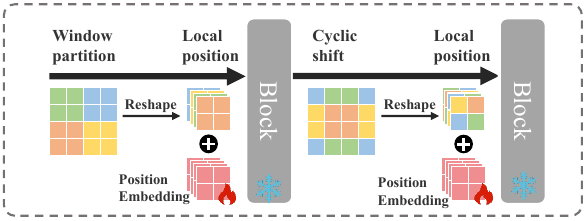}
    \vspace{-2em}
    \caption{Workflow of the \textit{shifted windows}. }
    \label{ShiftedWindows}
    \vspace{-1.5em}
\end{figure}

\begin{figure*}[t]
  \centering
    \includegraphics[width=1\linewidth]{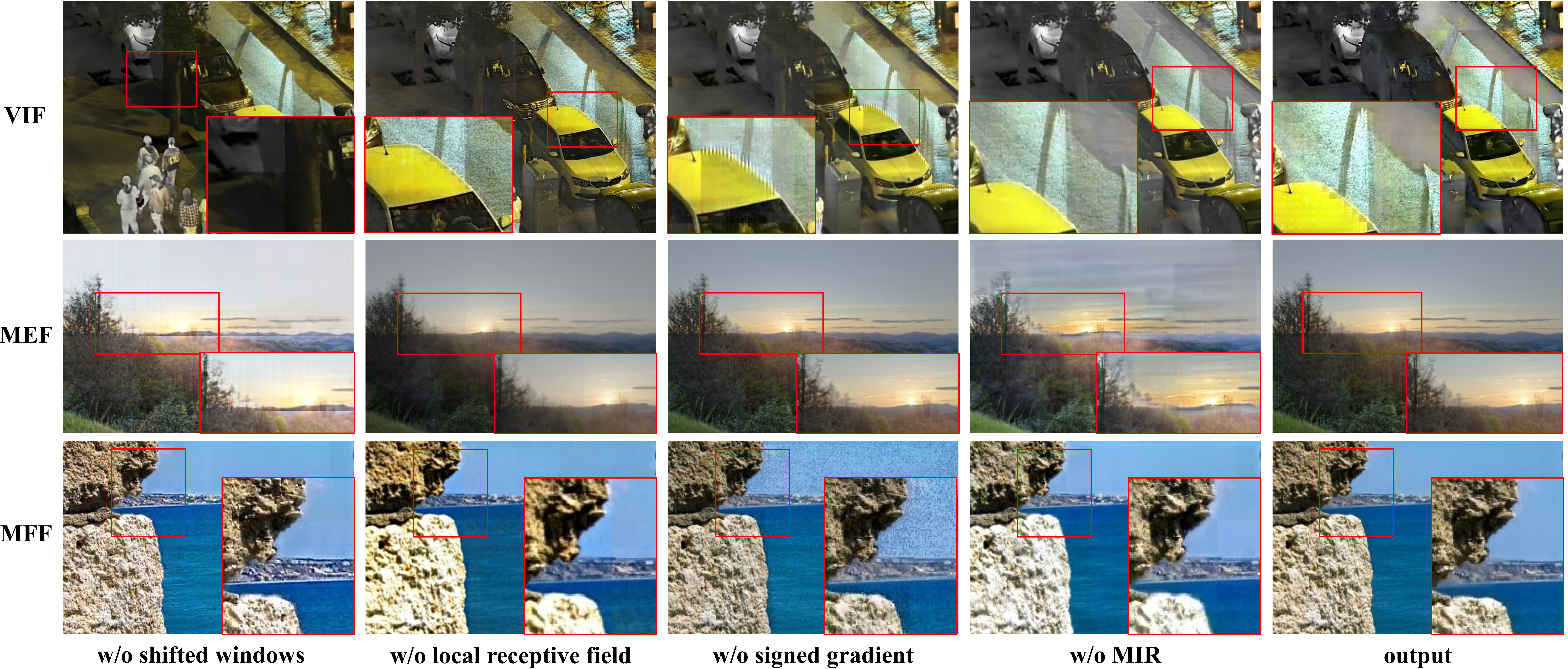}
    \vspace{-1.8em}
    \caption{Ablation studies about network design.}
    \label{Vis_network_design}
    \vspace{-0.6em}
\end{figure*}

\section{Details of Network.} \label{NetworkDetail}

\begin{figure*}[t]
  \centering
    \includegraphics[width=1\linewidth]{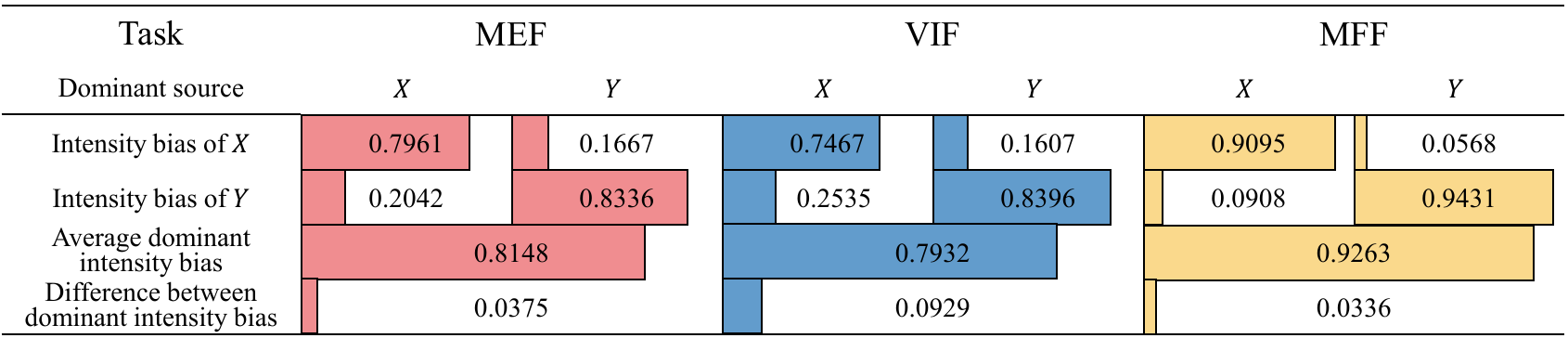}
    \vspace{-1.8em}
    \caption{Statistical analysis of intensity bias in different tasks. The “Average dominant intensity bias" refers to the average value of the dominant intensity bias for the $X$ and $Y$ sources. The “Difference between dominant intensity bias" denotes to the absolute value of the difference between the dominant intensity bias of the $X$ and $Y$ sources. }
    \label{IntensityAnalysis}
    \vspace{-1.3em}
\end{figure*}

\noindent\textbf{Detail Architecture of TC-MoA. } We illustrate the detailed network structure, data flow of the prompt generation and prompt-driven fusion stages of TC-MoA in \cref{DetailNetWork}. 

\noindent\textbf{Details on Fused Results. } With each passing TC-MoA module, the fusion features are added to each network branch according to hyperparameter settings. We consider that the features in each branch increasingly resemble those of the fused image, and ultimately the outputs of the two branches to be approximately identical, thus arbitrarily choosing one branch to obtain the fused image.

\noindent\textbf{Shifted Windows.}
We integrate windows shifting into the frozen ViT blocks by incorporating learnable relative position embeddings, as shown in~\cref{ShiftedWindows}. To reduce the computational cost of transformer, most previous approaches only support fixed-size inputs. These methods employ additional pre-processing and post-processing steps to crop and stitch the image together, resulting in the checkerboard artifacts. To address this issue, we partition the features into multiple windows of $ 14\times14 $ patches (to maintain consistency with the token length during pre-training). Then, we introduce learnable local position embeddings to enable the model's perception of token positions within the windows, ensuring the model is spatially aware of the local windows. Subsequently, we apply cyclic shifts to other blocks, acquiring a global receptive field. 
To this end, \textit{shifted windows} not only allows for efficiently handling of different input sizes, but also captures the global receptive field of the image to effectively avoid checkerboard artifacts.

\noindent\textbf{Ablation Studies about Network Design.}
We explore the effect of three network architectures in our framework. The qualitative results are shown in \cref{Vis_network_design}. 
i) Predicting images in isolated windows neglects the inter-window information exchange, thereby yielding inconsistencies across the entire image. Our method enables the interaction of information across the entire image by shifting windows, thereby maintaining the overall coherence of the image.
ii) The transformers are not able to effectively transmit information to adjacent patches solely based on long-range dependencies, resulting in the occurrence of the fusion image checkerboard effect. 
 To address this, we introduce local receptive fields via convolutional layers to enhance the interaction of local features and alleviate the checkerboard effect between adjacent patches. iii) Using only gradient constraints with absolute values can lead the network to blindly learn the gradients of the image, resulting in the generation of fusion images with incorrect gradient directions, which do not align with human visual perception. The loss function based on the signed value of the gradient avoids confusion in the direction of the gradient. 

%

\begin{table}[t]
\centering
\setlength{\tabcolsep}{5.5pt}
\begin{tabular}{cccc}
\hline
\rowcolor[HTML]{D9D9D9} 
\multicolumn{4}{c}{\cellcolor[HTML]{D9D9D9}All Parameters} \\
\multicolumn{4}{c}{348.7 M} \\ \hline
\rowcolor[HTML]{D9D9D9} 
\multicolumn{1}{c|}{\cellcolor[HTML]{D9D9D9}Frozen} & \multicolumn{3}{c}{\cellcolor[HTML]{D9D9D9}Trainable} \\
\multicolumn{1}{c|}{} & \multicolumn{3}{c}{9.58 M (2.82\%)} \\ \cline{2-4} 
\multicolumn{1}{c|}{} & \multicolumn{1}{c|}{\cellcolor[HTML]{D9D9D9}\begin{tabular}[c]{@{}c@{}}Position\\ Embedding\end{tabular}} & \multicolumn{2}{c}{\cellcolor[HTML]{D9D9D9}TC-MoA} \\
\multicolumn{1}{c|}{} & \multicolumn{1}{c|}{} & \multicolumn{2}{c}{9.39 M  (2.77\%)} \\ \cline{3-4} 
\multicolumn{1}{c|}{} & \multicolumn{1}{c|}{} & \cellcolor[HTML]{D9D9D9}\begin{tabular}[c]{@{}c@{}}Prompt\\ Generation\end{tabular} & \cellcolor[HTML]{D9D9D9}\begin{tabular}[c]{@{}c@{}}Prompt-Driven\\ Fusion\end{tabular} \\
\multicolumn{1}{c|}{\multirow{-5}{*}{\begin{tabular}[c]{@{}c@{}}329.54 M\\ (97.18\%)\end{tabular}}} & \multicolumn{1}{c|}{\multirow{-3}{*}{\begin{tabular}[c]{@{}c@{}}0.19 M\\ (0.06\%)\end{tabular}}} & \begin{tabular}[c]{@{}c@{}}2.15 M\\ (0.63\%)\end{tabular} & \begin{tabular}[c]{@{}c@{}}7.24 M\\ (2.13\%)\end{tabular} \\
\hline
\end{tabular}
\vspace{-0.5em}
 \caption{Detailed statistics of the parameters. The data is sourced from the TC-MoA model based on ViT-large network structure with $ \tau=4 $ and $ N=4 $.}
    \label{ParametersStat}
    \vspace{-1.8em}
\end{table}

\noindent\textbf{More Analysis of Parameters. } 
Our network is an efficient parameter fine-tuning method inserted into the frozen ViT framework with pre-trained parameters. As depicted in Table~\ref{ParametersStat}, we use mere 2.77~\% trainable parameters to bridge the gap between pre-trained model and image fusion tasks. The introduced shifted windows only demand 0.06 \% parameters, rendering them practically negligible.  In TC-MoA, although we have multiple routing networks and adapters, a mere 0.63\% parameters suffice to generate prompt efficiently. In fact, most of the parameters are contributed by the convolutional layers in the fusion layer. 
In addition, our model offers the potential to further reduce the number of parameters by compressing the number of channels in the convolutional layers.

\section{Additional Analysis and Discussion}\label{AdditionalAnalysis}

\noindent\textbf{The Properties of Various Fusion Tasks.}
\begin{table*}[t]
\centering
\footnotesize
\setlength{\tabcolsep}{6.2pt}
  \caption{ Quantitative results of the VIF task on  TNO dataset. }
  \vspace{-1em}
\scalebox{1.1}{
\begin{tabular}{@{}l|cccccccc@{}}
\toprule
Method & $\mathcal{VIF}$ & $ Q_c $ & EN & SD & \textbf{$Q_{cv}$↓} & MS-SSIM & FMI & $Q_w$ \\
\midrule
\rowcolor[HTML]{FFFFFF} 
DeFusion \cite{DeFusion}  {[}ECCV'22{]} & 0.513 & {{0.569}} & 6.579 & 8.862 & {{500.767}} & 0.840 & 0.900 & 0.563 \\
\rowcolor[HTML]{FFFFFF} 
DDFM \cite{DDFM}  {[}ICCV'23{]} & 0.276 & 0.390 & 6.853 &  {{9.219}} & 976.884 & 0.685 & 0.878 & 0.294 \\
\rowcolor[HTML]{FFFFFF} 
MoE-Fusion \cite{MoE-Fusion} {[}ICCV'23{]} & {\color[HTML]{0000FF} \textbf{0.757}} & 0.541 & {\color[HTML]{0000FF} \textbf{7.008}} & 9.158 & 743.774 & {\color[HTML]{0000FF} \textbf{0.901}} &  {{0.907}} &  {{0.770}} \\
\rowcolor[HTML]{FFFFFF} 
TC-MoA \textit{Base} &  {{0.748}} & {\color[HTML]{0000FF} \textbf{0.631}} &  {{7.003}} & {\color[HTML]{0000FF} \textbf{9.335}} & {\color[HTML]{FF0000} \textbf{393.571}} &  {{0.895}} & {\color[HTML]{FF0000} \textbf{0.911}} & {\color[HTML]{0000FF} \textbf{0.770}} \\
\rowcolor[HTML]{FFFFFF} 
TC-MoA \textit{Large} & {\color[HTML]{FF0000} \textbf{0.793}} & {\color[HTML]{FF0000} \textbf{0.659}} & {\color[HTML]{FF0000} \textbf{7.026}} & {\color[HTML]{FF0000} \textbf{9.392}} & {\color[HTML]{0000FF} \textbf{414.068}} & {\color[HTML]{FF0000} \textbf{0.908}} & {\color[HTML]{0000FF} \textbf{0.910}} & {\color[HTML]{FF0000} \textbf{0.772}} 
 \\
\bottomrule 
\end{tabular} }
\label{VIF_TNO_Exp}%
\vspace{-1em}
\end{table*}
\begin{table*}[t]
\centering
\footnotesize
\setlength{\tabcolsep}{7.2pt}
  \caption{ Ablation experiments
on the value of $K$. }
  \vspace{-1em}
\scalebox{1.1}{
\begin{tabular}{@{}c|ccc|ccc|ccc}
\toprule
 & \multicolumn{3}{c|}{VIF} & \multicolumn{3}{c|}{MEF} & \multicolumn{3}{c}{MFF} \\
\multirow{-2}{*}{Hyperparameter} & \cellcolor[HTML]{EFEFEF}$Q_{abf}$ & \cellcolor[HTML]{EFEFEF}$Q_{p}$ & \cellcolor[HTML]{EFEFEF}SSIM & \cellcolor[HTML]{EFEFEF}$Q_{abf}$ & \cellcolor[HTML]{EFEFEF}$Q_{p}$ & \cellcolor[HTML]{EFEFEF}SSIM & \cellcolor[HTML]{EFEFEF}$Q_{abf}$ & \cellcolor[HTML]{EFEFEF}$Q_{p}$ & \cellcolor[HTML]{EFEFEF}SSIM \\ \midrule
$K=1$ & 0.608 & 0.731 & 0.456 & 0.655 & 0.897 & {\color[HTML]{0000FF} \textbf{0.679}} & {\color[HTML]{0000FF} \textbf{0.652}} & {\color[HTML]{FF0000} \textbf{0.674}} & 0.413 \\
$K=2$& 0.608 & {\color[HTML]{FF0000} \textbf{0.739}} & {\color[HTML]{FF0000} \textbf{0.458}} & 0.656 & 0.896 & 0.678 & 0.649 & 0.665 & 0.411 \\
$K=3$ & {\color[HTML]{FF0000} \textbf{0.610}} & 0.734 & {\color[HTML]{0000FF} \textbf{0.457}} & {\color[HTML]{0000FF} \textbf{0.656}} & {\color[HTML]{FF0000} \textbf{0.898}} & 0.679 & 0.652 & 0.669 & {\color[HTML]{FF0000} \textbf{0.413}} \\
$K=4$ & {\color[HTML]{0000FF} \textbf{0.609}} & {\color[HTML]{0000FF} \textbf{0.736}} & 0.457 & {\color[HTML]{FF0000} \textbf{0.658}} & {\color[HTML]{0000FF} \textbf{0.897}} & {\color[HTML]{FF0000} \textbf{0.680}} & {\color[HTML]{FF0000} \textbf{0.653}} & {\color[HTML]{0000FF} \textbf{0.669}} & {\color[HTML]{0000FF} \textbf{0.413}} \\
\bottomrule
\end{tabular}}
\label{suppl_topK}%
\vspace{-1em}
\end{table*}
The properties of a fusion task, as well as the relationship between multiple tasks, can be depicted through the task prompt. For each token pair, if $prompt_{x} > prompt_{y}$, then we refer to $X$ as the dominant source and $Y$ as the auxiliary source. The dominant intensity bias of $X$ is defined as the average value of $prompt_{x}$ for all token pairs where $X$ dominates. Conversely, the auxiliary intensity bias of $X$ is the mean value of $prompt_{x}$ for all token pairs where $X$ is the auxiliary source. As shown in~\cref{IntensityAnalysis}, for an instance, the dominant and auxiliary intensity biases of $X$ for the VIF task are 0.7467 and 0.1607, respectively.


\cref{IntensityAnalysis} reveals two fusion patterns: 1) The mean dominant intensity bias for MFF is higher than those of MEF and VIF. This indicates that fused images from MFF tasks typically draw information heavily from one source image per token pair, rendering the fusion for MFF extremely unbalanced at the token-level. In contrast, MEF and VIF display a more balanced fusion. From a token-level fusion perspective, MEF and VIF tasks share more similarity.
2) The difference between dominant intensity bias of VIF is significantly larger than that of the other fusion tasks. For the VIF task, when the infrared images ($Y$ source) dominate, the dominant intensity bias is found to greatly surpass that of the visible images ($X$ source). Once the network identifies the dominance of infrared images, it assigns higher weights to its features. This results in VIF being more unbalanced at the source-level, while MEF and MFF exhibit more balance. We believe that part of the reason for this phenomenon is that the LLVIP dataset for the VIF task, composed primarily of nocturnal scenes, where infrared images might provide more information than visible images.

To this end, the average dominant intensity bias reflects the fusion pattern of the tasks at the token-level, while the difference between dominant intensity bias illustrates the pattern at the source-level. 
The empirical evidence of our method shows variations and connections among these fusion tasks, thus demonstrating our effectiveness in handling multiple fusion tasks with a unified model.
\begin{figure*}[t]
  \centering
 
    \includegraphics[width=0.9\linewidth]{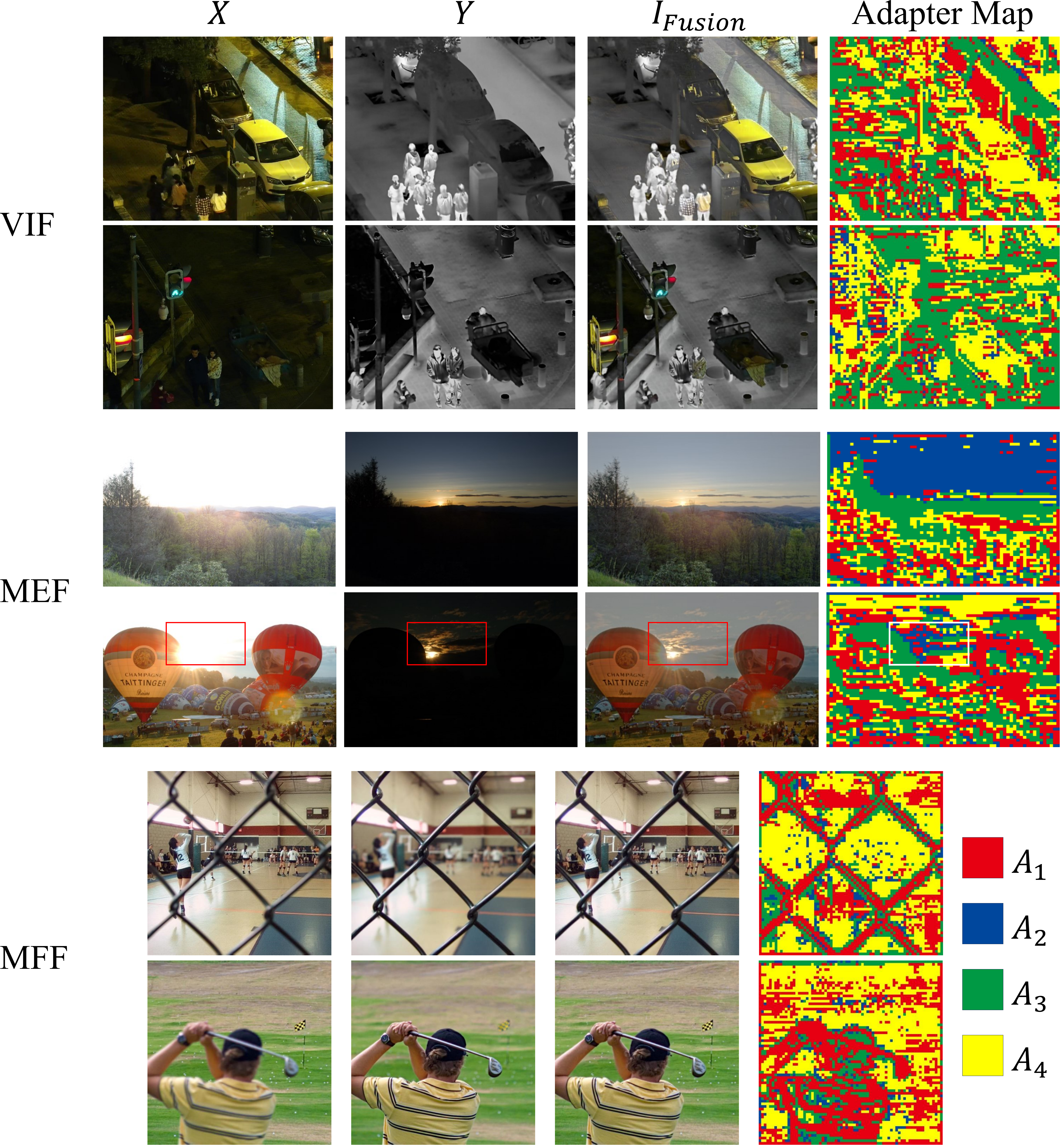}
    \caption{The visualization of adapter selection for the last TC-MoA in the ViT encoder. We visualize the index of the adapters with the highest routing weight as the adapter map. Different colors in the adapter map represent the selection of different adapters. It is worth noting that this is the adapters selection situation of the last TC-MoA in the encoder, which can only reflect the general trend and may contain unclear noise. }

    \label{AdapterMap}
\end{figure*}

\noindent\textbf{Details of Task-Specific Routing.}
Different task-specific routers indeed customized various mixtures of adapters, as shown in~\cref{AdapterMap}. The following patterns can be observed: 1) For the VIF task, if source $X$ is the dominant source on one token pair, it is inclined to utilize the yellow adapter for processing. Conversely, the green adapter is primarily used when source $Y$ predominates.  2) For the MEF task, the blue adapter is used when source $Y$ is the dominant source. Other colored adapters are utilized under conditions where source $X$ is dominant. 3) For the MFF task, the yellow adapter tends to deal with cases where $X$ dominates, while red and green tend to address situations where $Y$ is dominant. Notably, high-frequency areas are more frequently handled by the red adapter, while low-frequency areas are inclined to handle by the green adapter.

Obviously, by task-specific routing, the network tends to select different mixtures of adapters to accommodate varying tasks. Therefore, these adapters have task tendencies and different divisions of labor (such as high and low frequency works). This is an interesting finding that can be further explored for more controllable fusions.

\begin{figure*}[t]
  \centering
    \includegraphics[width=1\linewidth]{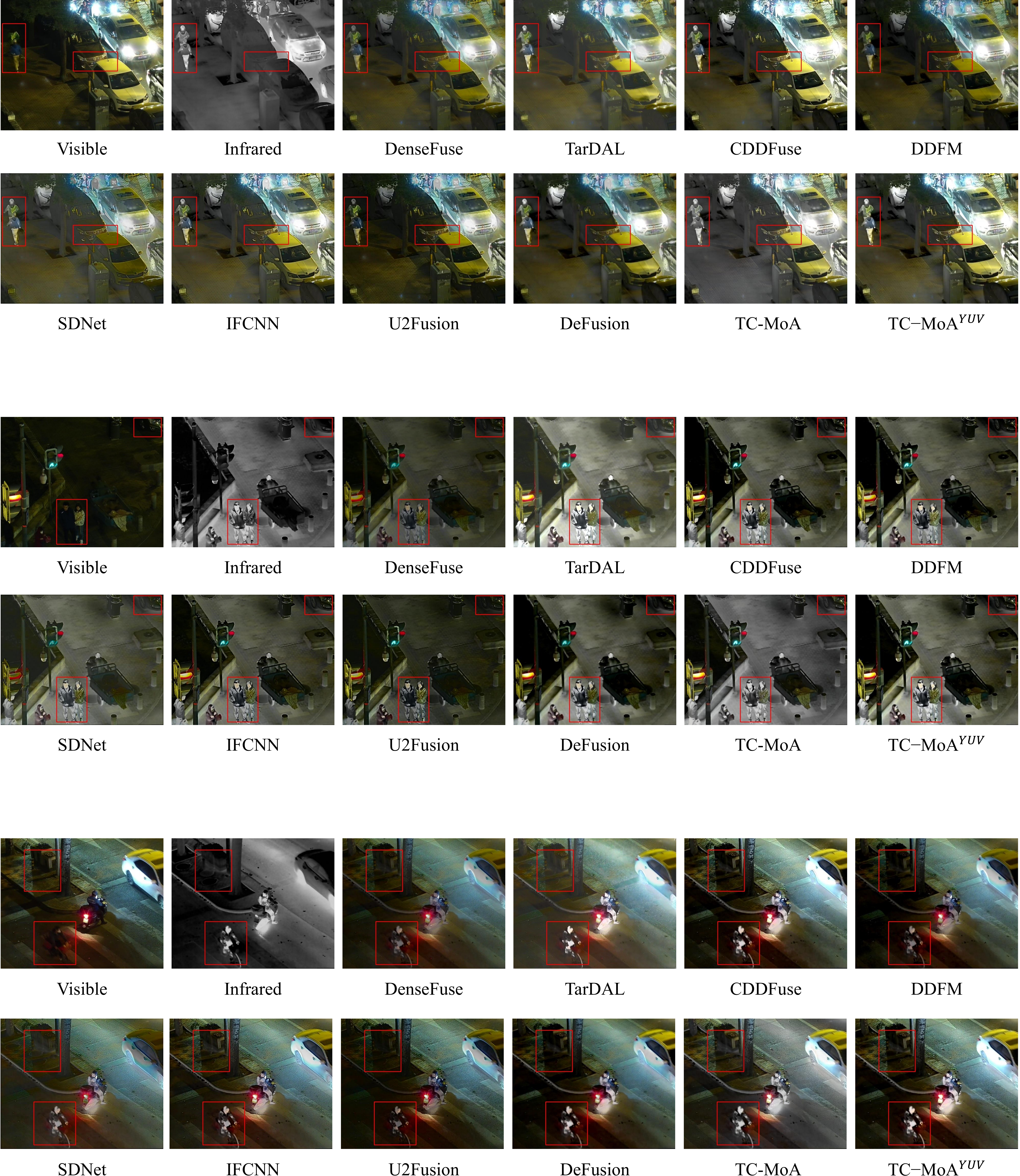}
    \caption{More qualitative comparisons in the VIF task. }
    \label{VIF_suppl}
    \vspace{1.5em}
\end{figure*}
\begin{figure*}[t]
  \centering
    \includegraphics[width=1\linewidth]{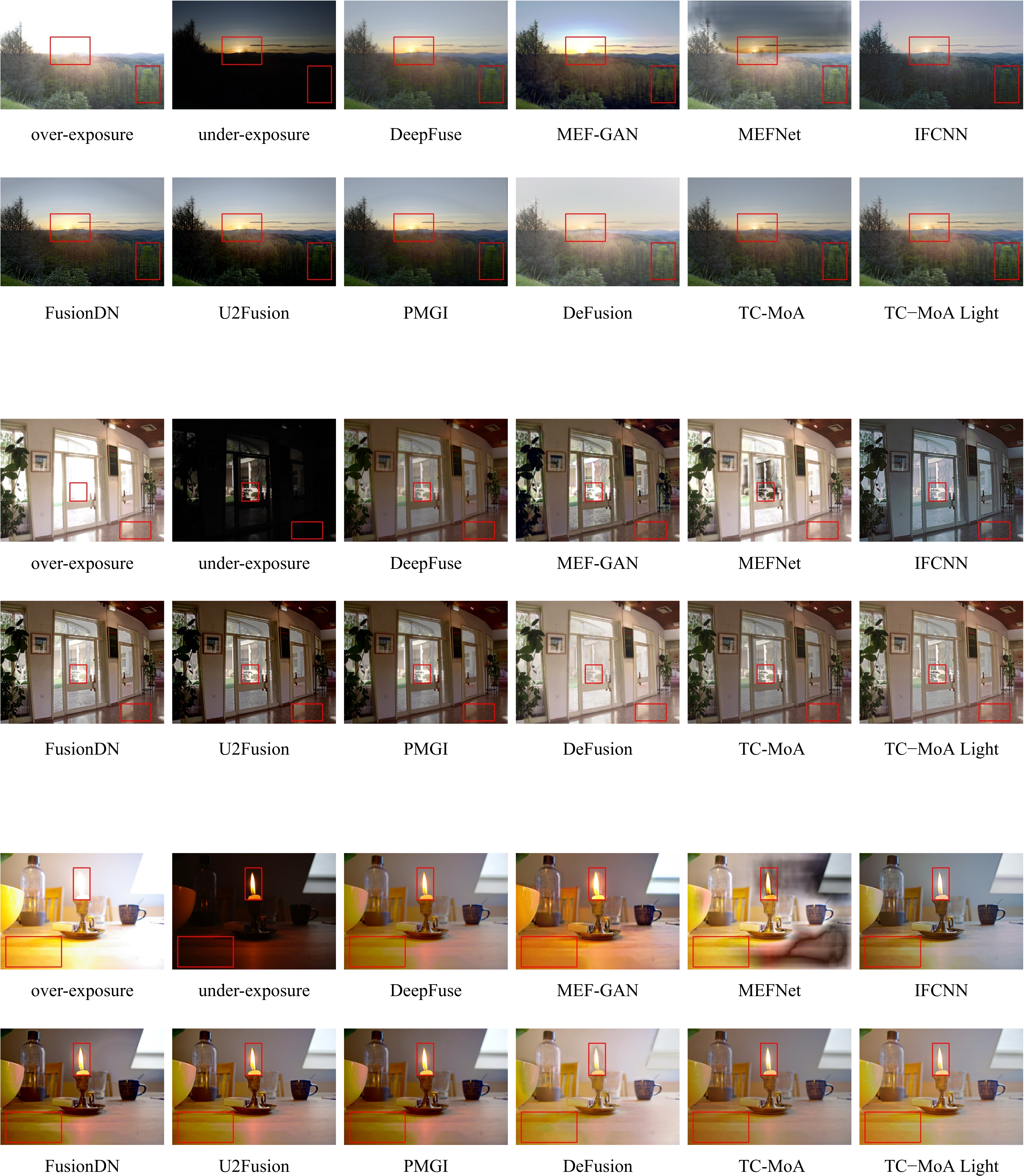}
    \caption{More qualitative comparisons in the MEF task. }
    \label{MEF_suppl}
    \vspace{1.5em}
\end{figure*}

\begin{figure*}[t]
  \centering
    \includegraphics[width=0.8\linewidth]{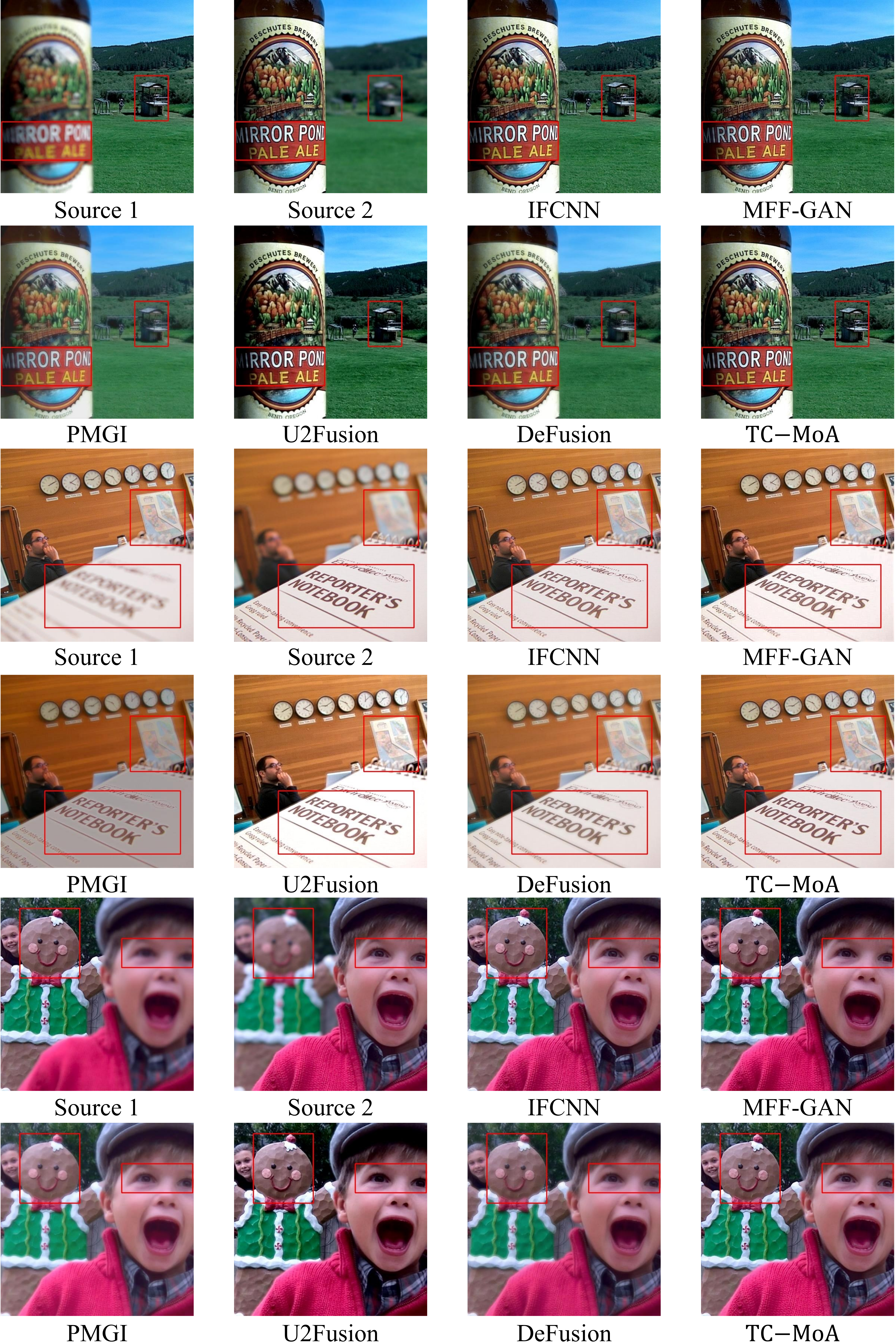}
    \caption{More qualitative comparisons in the MFF task. }
    \label{MFF_suppl}
    
\end{figure*}

\noindent\textbf{Analysis on Top $ K $.} 
We performed ablation experiments on the value of $ K $ under four adapters, as shown in Table~\ref{suppl_topK}. The following two findings can be observed: i) the more the number of experts chosen for routing, the better the overall performance. Intuitively, the more adapters are routed to, the greater the dynamism of the network, and consequently the better the performance, which is consistent with empirical results. ii) The choice of $K$ has a minor impact on performance, indicating that the network is not sensitive to this parameter. In summary, we have struck a balance between performance and inference cost by setting Top $K = 2$.

\section{More Quantitative Comparisons}\label{Morequantitative}
\textbf{TNO Dataset.} Table~\ref{VIF_TNO_Exp} shows additional results on TNO dataset. Our method achieves superior overall performance compared with the most recent methods.

\section{More Qualitative Comparisons} \label{Morequalitative}
More qualitative results of various fusion tasks are presented in~\cref{VIF_suppl,MEF_suppl,MFF_suppl}. In the case of the VIF task, our fusion results maintain the detailed information of the visible images to the greatest extent (such as license plate numbers), while clearly highlighting the information from the infrared images. The overall images possess excellent brightness and contrast. For the MEF task, our fusion results adequately preserve the detailed structural and color information from both sources, especially avoiding blurry halos in overexposed areas. In terms of the MFF task, our model avoids distortions in structural details and color, particularly in the areas with text. In terms of the overall visual perception, our method is comparable to the IFCNN on MFF, a supervised training method. In addition, our method provides a substantial degree of fusion controllability, making it practicable to manipulate fusion results based on particular requirements.